\documentclass[10pt,twocolumn,letterpaper]{article}

\usepackage{cvpr}
\usepackage{times}
\usepackage{epsfig}
\usepackage{graphicx}
\usepackage{amsmath}
\usepackage{amssymb}
\usepackage{enumitem}
\usepackage{subfigure}

\usepackage[pagebackref=true,breaklinks=true,letterpaper=true,colorlinks,bookmarks=false]{hyperref}

\cvprfinalcopy

\def\cvprPaperID{3333} 

\begin{document}

\title{FineGAN: Unsupervised Hierarchical Disentanglement for\\Fine-Grained  Object Generation and Discovery}

\author{Krishna Kumar Singh\thanks{Equal contribution.} ~~~~~~ Utkarsh Ojha\footnotemark[1] ~~~~~~ Yong Jae Lee \\
University of California, Davis
}

\maketitle

\begin{abstract}
We propose FineGAN, a novel unsupervised GAN framework, which disentangles the background, object shape, and object appearance to hierarchically generate images of fine-grained object categories. To disentangle the factors without supervision, our key idea is to use information theory to associate each factor to a latent code, and to condition the relationships between the codes in a specific way to induce the desired hierarchy. Through extensive experiments, we show that FineGAN achieves the desired disentanglement to generate realistic and diverse images belonging to fine-grained classes of birds, dogs, and cars.  Using FineGAN's automatically learned features, we also cluster real images as a first attempt at solving the novel problem of unsupervised fine-grained object category discovery. Our code/models/demo can be found at \url{https://github.com/kkanshul/finegan} 
\end{abstract}

\section{Introduction}

\vspace{-7pt}
\begin{figure}[h!]
    \centering
    \includegraphics[width=0.48\textwidth]{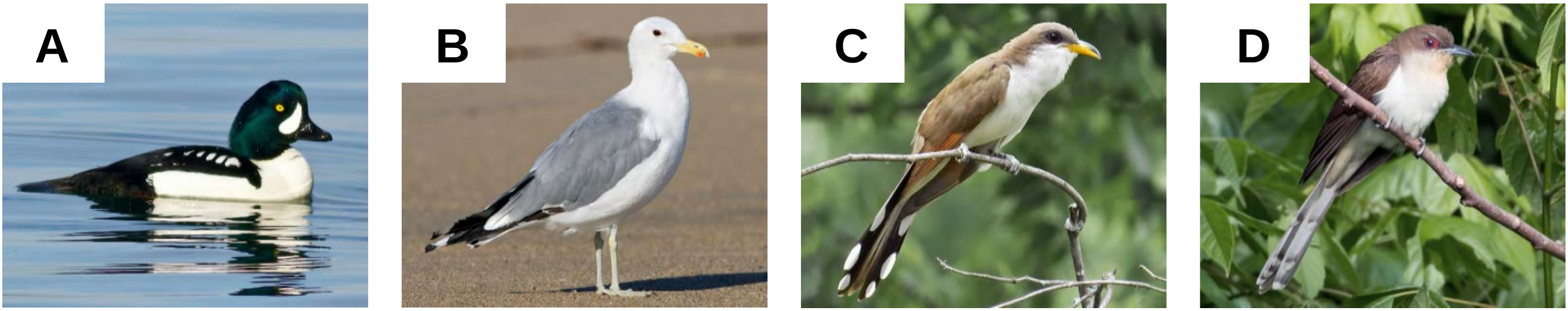}
    \label{fig:motivation}
\end{figure}
\vspace{-20pt}

Consider the figure above: if tasked to group any of the images together, as humans we can easily tell that birds A and B should not be grouped with C and D as they have completely different backgrounds and shapes.  But how about C and D?  They share the same background, shape, and rough color. However, upon close inspection, we see that even C and D should not be grouped together as C's beak is yellow and its tails have large white spots while D's beak is black and its tails have thin white strips.\footnote{The ground-truth fine-grained categories are A: \emph{Barrow's Goldeneye}, B: \emph{California Gull}, C: \emph{Yellow-billed Cuckoo}, D: \emph{Black-billed Cuckoo}.}  This example demonstrates that clustering fine-grained object categories requires not only \emph{disentanglement} of the background, shape, and appearance (color/texture), but that it is naturally facilitated in a hierarchical fashion.

\begin{figure}[t!]
    \centering
    \includegraphics[width=0.48\textwidth]{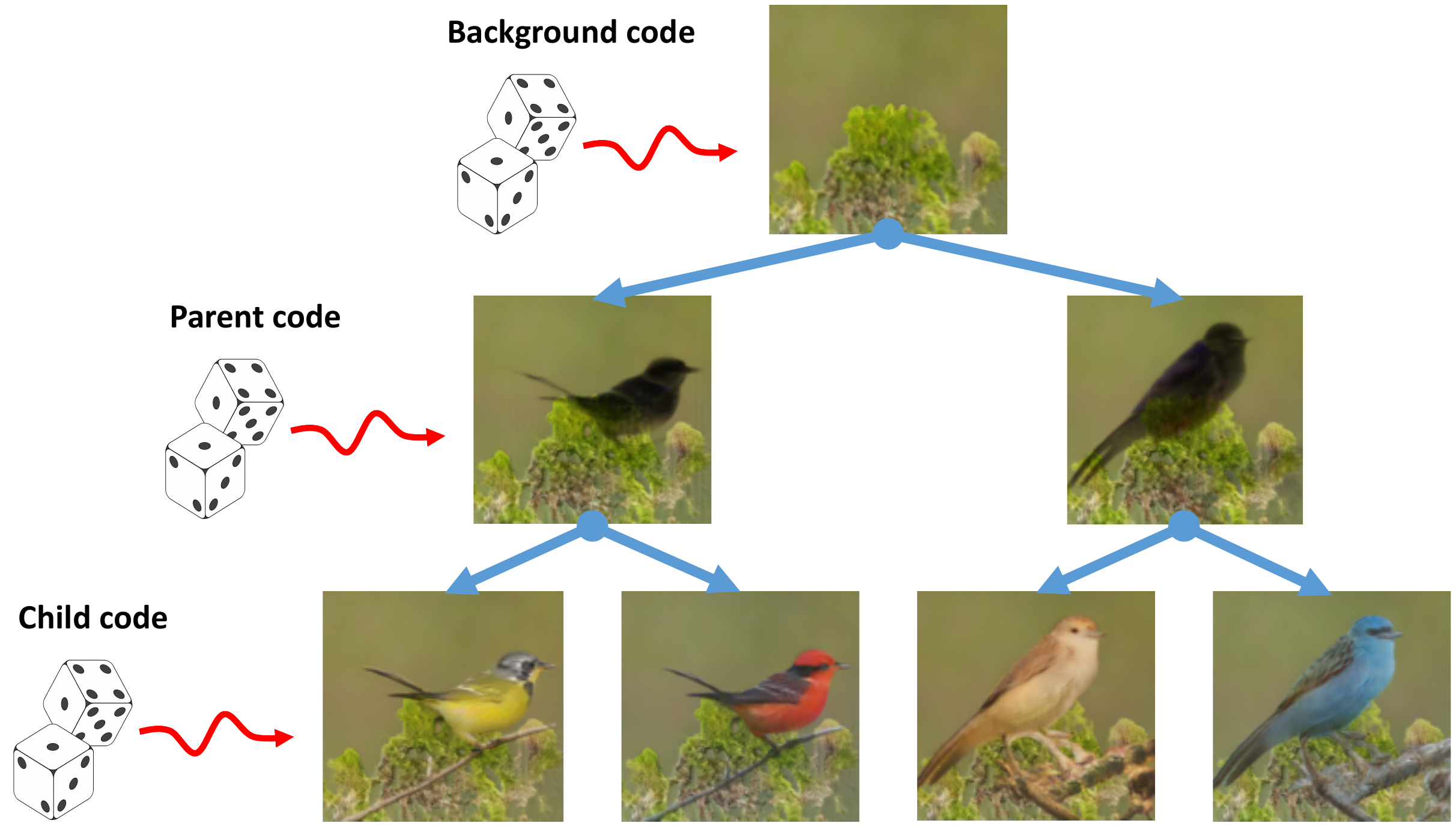}
    \caption{FineGAN disentangles the background, object shape (parent), and object appearance (child) to hierarchically generate fine-grained objects, \emph{without mask or fine-grained annotations}.}
    \label{fig:teaser}
\end{figure}

In this work, we aim to develop a model that can do just that: model fine-grained object categories by hierarchically disentangling the background, object's shape, and its appearance, without any manual fine-grained annotations. Specifically, we make the first attempt at solving the novel problem of \emph{unsupervised} fine-grained object clustering (or ``discovery'').  Although both unsupervised object discovery and fine-grained recognition have a long history, prior work on unsupervised object category discovery focus only on clustering entry-level categories (e.g., birds vs.~cars vs.~dogs)~\cite{grauman-cvpr06,sivic-cvpr08,lee-cvpr10,yang-cvpr16,xie-icml16,dizaji-iccv17}, while existing work on fine-grained recognition focus exclusively on the supervised setting in which ground-truth fine-grained category annotations are provided~\cite{nilsback-icvgip08,yao-cvpr11,liu-eccv12,berg-cvpr13,gavves-iccv13,lin-iccv15,gao-cvpr16,cai-iccv17,wang-cvpr18}.

Why unsupervised discovery for such a difficult problem?  We have two key motivations.  First, fine-grained annotations require domain experts. As a result, the overall annotation process is very expensive and standard crowdsourcing techniques cannot be used, which restrict the amount of training data that can be collected.  Second, unsupervised learning enables the discovery of \emph{latent structure} in the data, which may not have been labeled by annotators.  For example, fine-grained image datasets often have an inherent \emph{hierarchical organization} in which the categories can first be grouped based on one feature (e.g., shape) and then differentiated based on another (e.g., appearance).

\vspace{-10pt}
\paragraph{Main Idea.} We hypothesize that a generative model with the capability of hierarchically generating images with fine-grained details can also be useful for fine-grained grouping of real images.  We therefore propose \emph{FineGAN}, a novel hierarchical unsupervised Generative Adversarial Networks framework to generate images of fine-grained categories.

FineGAN generates a fine-grained image by hierarchically generating and stitching together a background image, a parent image capturing one factor of variation of the object, and a child image capturing another factor. To disentangle the two factors of variation of the object \emph{without any supervision}, we use information theory, similar to InfoGAN ~\cite{chen-nips16}. Specifically, we enforce high mutual information between (1) the parent latent code and the parent image, and (2) the child latent code, \emph{conditioned on the parent code}, and the child image.  By imposing constraints on the relationship between the parent and child latent codes (specifically, by grouping child codes such that each group has the same parent code), we can induce the parent and child codes to capture the object's shape and color/texture details, respectively; see Fig.~\ref{fig:teaser}.  This is because in many fine-grained datasets, objects often differ in appearance conditioned on a shared shape (e.g., `Yellow-billed Cuckoo' and `Black-billed Cuckoo', which share the same shape but differ in their beak color and wing patterns).

Moreover, FineGAN automatically generates masks at both the parent and child stages, which help condition the latent codes to focus on the relevant object factors as well as to stitch together the generated images across the stages.  Ultimately, the features learned through this unsupervised hierarchical image generation process can be used to cluster real images into their fine-grained classes.

\vspace{-10pt}
\paragraph{Contributions.} Our work has two main contributions:

(1) We introduce FineGAN, an unsupervised model that learns to hierarchically generate the background, shape, and appearance of fine-grained object categories. Through various qualitative evaluations, we demonstrate FineGAN's ability to accurately disentangle background, object shape, and object appearance. Furthermore, quantitative evaluations on three benchmark datasets (CUB~\cite{wah-tech11}, Stanford-dogs~\cite{khosla-FGVC11}, and Stanford-cars~\cite{krause-DRR2013}) demonstrate FineGAN's strength in generating realistic and diverse images.

(2) We use FineGAN's learned disentangled representation to cluster real images for unsupervised fine-grained object category discovery.  It produces fine-grained clusters that are significantly more accurate than those of state-of-the-art unsupervised clustering approaches (JULE~\cite{yang-cvpr16} and DEPICT~\cite{dizaji-iccv17}).  To our knowledge, this is the first attempt to cluster fine-grained categories in the unsupervised setting.  

 \section{Related work}

\paragraph{Fine-grained category recognition} involves classifying subordinate categories within entry-level categories (e.g., different species of birds), which requires annotations from domain experts~\cite{nilsback-icvgip08,yao-cvpr11,liu-eccv12,berg-cvpr13,gavves-iccv13,chai-iccv13,lin-iccv15,gao-cvpr16,kong-cvpr17,zheng-iccv17,wang-cvpr18}. Some methods require additional part~\cite{zhang-eccv14,branson-bmvc14,zhang-cvpr16}, attribute~\cite{gebru-iccv17}, or text~\cite{reed-cvpr16,he-cvpr17} annotations.  Our work makes the first attempt to overcome the dependency on expert annotations by performing \emph{unsupervised} fine-grained category discovery without any class annotations.  
\vspace{-10pt}
\paragraph{Visual object discovery and clustering.}
Early work on unsupervised object discovery~\cite{sivic-iccv05,grauman-cvpr06,sivic-cvpr08,lee-cvpr10,lee-cvpr11,rubinstein-cvpr13} use handcrafted features to cluster object categories from unlabeled images. Others explore the use of natural language dialogue for object discovery~\cite{vries-cvpr17,zhuang-cvpr18}. Recent unsupervised deep clustering approaches~\cite{yang-cvpr16,xie-icml16,dizaji-iccv17} demonstrate state-of-the-art results on datasets whose objects have large variations in high-level detail like shape and background. On fine-grained category datasets, we show that FineGAN significantly outperforms these methods as it is able to focus on the fine-grained object details.

\vspace{-10pt}
\paragraph{Disentangled representation learning} has a vast literature (e.g.,~\cite{bengio-tpami2013,tenenbaum-2000,hinton-icann2011,yan-eccv16,chen-nips16,higgins-iclr2017,denton-nips17,hu-cvpr18}).  The most related work in this space is InfoGAN~\cite{chen-nips16}, which learns disentangled representations without any supervision by maximizing the mutual information between the latent codes and generated data.  Our work builds on the same principles of information theory, but we extend it to learn a \emph{hierarchical} disentangled representation. Specifically, unlike InfoGAN in which all details of an object are generated together, FineGAN provides explicit distentanglement and control over the generation of background, shape, and appearance, which we show is especially important when modeling fine-grained categories.

\vspace{-10pt}
\paragraph{GANs and Stagewise image generation.}
Unconditional GANs~\cite{goodfellow-nips2014,radford-iclr16,taigman-iclr17,zhao-iclr17,arjovsky-arxiv17,gulrajani-nips17} can generate realistic images without any supervision. However, unlike our approach, these methods do not generate images hierarchically and do not have explicit control over the background, object's shape, and object's appearance.  Some conditional supervised approaches~\cite{reed-icml2016,zhang-iccv17,stackgan2,bodla-eccv18} learn to generate fine-grained images with text descriptions.   One such approach, FusedGAN~\cite{bodla-eccv18}, generates fine-grained objects with specific pose and shape but it cannot decouple them, and lacks explicit control over the background. In contrast, FineGAN can generate fine-grained images without any text supervision and with full control over the background, pose, shape, and appearance.  Also related are stagewise image generators~\cite{im-arxiv16,kwak-arxiv16,yang-iclr17,johnson-cvpr18}. In particular, LR-GAN~\cite{yang-iclr17} generates the background and foreground separately and stitches them. However, both are controlled by a single random vector, and it does not disentangle the object's shape from appearance.  

\section{Approach}

Let $\mathcal{X} = \{x_1, x_2,\dots,x_N\}$ be a dataset containing unlabeled images of fine-grained object categories. Our goal is to learn an unsupervised generative model, FineGAN, which  produces high quality images matching the true data distribution $p_{data}(x)$, while also learning to disentangle the relevant factors of variation associated with images in $\mathcal{X}$.

We consider background, shape, appearance, and pose/location of the object as the factors of variation in this work. If FineGAN can successfully associate each latent code to a particular fine-grained category aspect (e.g., like a bird's shape and wing color), then its learned features can also be used to group the real images in $\mathcal{X}$ for unsupervised find-grained object category discovery.

\begin{figure*}[t!]
    \centering
    \includegraphics[width=0.95\textwidth]{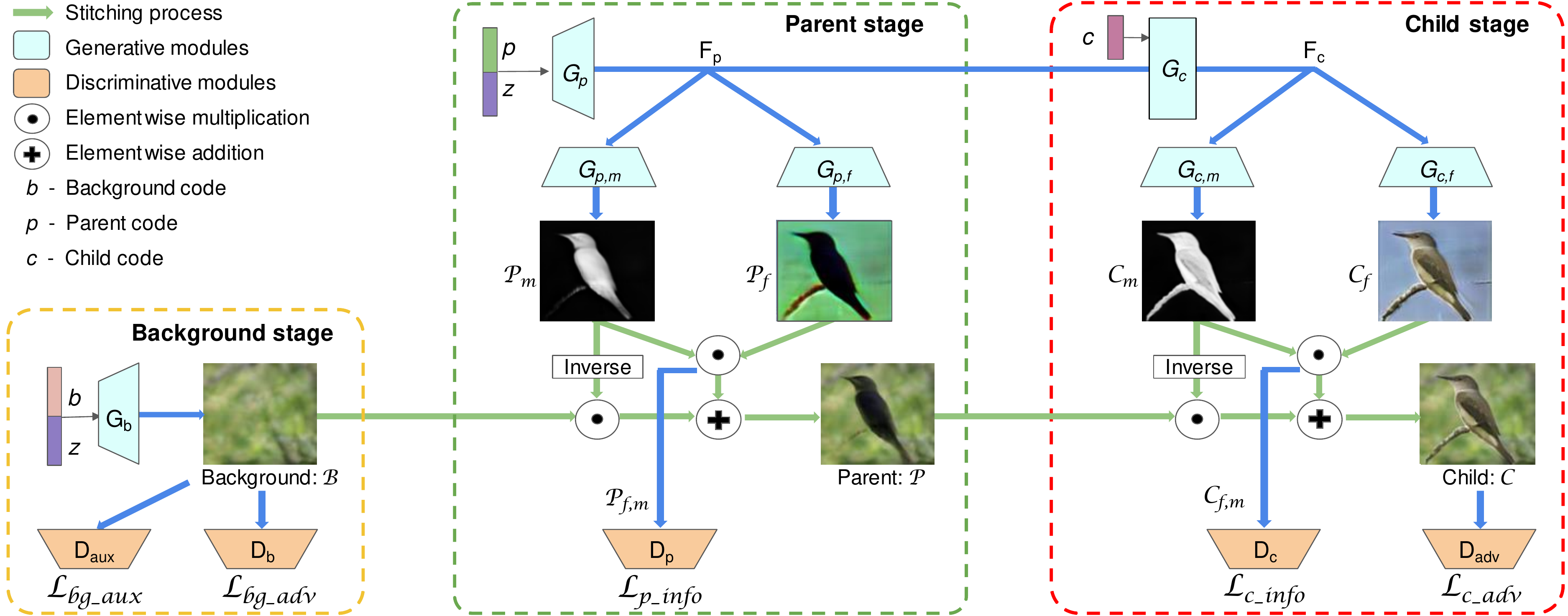}
    \caption{\textbf{FineGAN architecture} for hierarchical fine-grained image generation. The background stage, conditioned on random vector $z$ and background code $b$, generates the background image $B$. The parent stage, conditioned on $z$ and parent code $p$, uses $B$ as a canvas to generate parent image $P$, which captures the shape of the object. The child stage, conditioned on $c$, uses $P$ as a canvas to generate the final child image $C$ with the object's appearance details stitched into the shape outline.}
    \label{fig:main}
    \vspace{-0.09in}
\end{figure*}

\subsection{Hierarchical fine-grained disentanglement}\label{rel}

Fig.~\ref{fig:main} shows our FineGAN architecture for modeling and generating fine-grained object images. The overall process has three interacting stages: background, parent, and child. The background stage generates a realistic background image $\mathcal{B}$.  The parent stage generates the outline (shape) of the object and stitches it onto $\mathcal{B}$ to produce parent image $\mathcal{P}$. The child stage fills in the object's outline with the appropriate color and texture, to produce the final child image $\mathcal{C}$.  The objective function of the complete process is:
\begin{equation*}\label{eq1}
\mathcal{L} = \lambda \mathcal{L}_{b} + \beta \mathcal{L}_{p} + \gamma \mathcal{L}_{c}
\end{equation*}
where $\mathcal{L}_{b}$, $\mathcal{L}_{p}$ and $\mathcal{L}_{c}$ denote the objectives for the background, parent, and child stage respectively, with $\lambda$, $\beta$ and $\gamma$ denoting their weights.  We train all stages end-to-end.

The different stages get conditioned with different latent codes, as seen from Fig.~\ref{fig:main}. FineGAN takes as input: i) a continuous noise vector $z \sim \mathcal{N} (0,1)$; ii) a categorical background code $b \sim \text{Cat}(K = N_b,p = 1/N_b)$; iii) a categorical parent code $p \sim \text{Cat}(K = N_p,p = 1/N_p)$; and iv) a categorical child code $c \sim \text{Cat}(K = N_c,p = 1/N_c)$.

\vspace{-5pt}
\paragraph{Relationship between latent codes:}

(1) \emph{Parent code and child code.} We assume the presence of an implicit hierarchy in $\mathcal{X}$ -- as mentioned previously, fine-grained categories can often be grouped first based on a common shape and then differentiated based on appearance. To help discover this hierarchy, we impose two constraints: (i) the number of categories of parent code is set to be less than that of child code ($N_p < N_c$), and (ii) for each parent code $p$, we tie a fixed number of child codes $c$ to it (multiple child codes share the same parent code).  We will show that these constraints help push $p$ to capture shape and $c$ to capture appearance. For example, if the shape identity captured by $p$ is that of a duck, then the list of $c$'s tied to this $p$ would all share the same duck shape, but vary in their color and texture.

(2) \emph{Background code and child code.} There is usually some correlation between an object and the background in which it is found (e.g., ducks in water). Thus, to avoid conflicting object-background pairs (which a real/fake discriminator could easily exploit to tell that an image is fake), we set the background code to be the same as the child code during training ($b = c$). However, we can easily relax this constraint during testing (e.g., to generate a duck in a tree).

\vspace{-5pt}
\subsubsection{Background stage}

The background stage synthesizes a background image $\mathcal{B}$, which acts as a canvas for the parent and child stages to stitch different foreground aspects on top of $\mathcal{B}$. Since we aim to disentangle background as a separate factor of variation, $\mathcal{B}$ should not contain any foreground information. We hence separate the background stage from the parent and child stages, which share a common feature pipeline.  This stage consists of a generator $G_{b}$ and a discriminator pair, $D_{b}$ and $D_{aux}$. $G_{b}$ is conditioned on latent background code $b$, which controls the different (unknown) background classes (e.g., trees, water, sky), and on latent code $z$, which controls intra-class background details (e.g., positioning of leaves).
To generate the background, we assume access to an object bounding box detector that can detect instances of the super-category (e.g., bird).  We use the detector to locate non-object background patches in each real image $x_i$.  We then train $G_b$ and $D_b$ using two objectives: $\mathcal{L}_{b} = \mathcal{L}_{bg\textunderscore adv} + \mathcal{L}_{bg\textunderscore aux}$, where $\mathcal{L}_{bg\textunderscore adv}$ is the adversarial loss~\cite{goodfellow-nips2014} and $\mathcal{L}_{bg\textunderscore aux}$ is the auxiliary background classification loss.

For the adversarial loss $\mathcal{L}_{bg\textunderscore adv}$, we employ the discriminator $D_b$ on a patch level~\cite{isola-cvpr17} (we assume background can easily be modeled as texture) to predict an $N \times N$ grid with each member indicating the real/fake score for the corresponding patch in the input image:
\vspace{-2pt}
\begin{equation*}
\resizebox{1.1 \columnwidth}{!} {\resizebox{\hsize}{!}{
    $\mathcal{L}_{bg\textunderscore adv} = \min\limits_{G_b}\max\limits_{D_b}~\mathbb{E}_{x}[\log (D_b(x))]
+ \mathbb{E}_{z, b}[\log (1 - D_b(G_b(z, b)))]$ 
}}\vspace{-3pt}
\label{eq2}
\end{equation*}

The auxiliary classification loss $\mathcal{L}_{bg\textunderscore aux}$ makes the background generation task more explicit, and is also computed on a patch level. Specifically, patches inside ($r_{i}$) and outside ($r_{o}$) the detected object in real images constitute the training set for foreground (1) and background (0) respectively, and is used to train a binary classifier $D_{aux}$ with cross-entropy loss.  We then use $D_{aux}$ to train the generator $G_b$:
\vspace{-2pt}
\begin{align*}
\mathcal{L}_{bg\textunderscore aux} = \min\limits_{G_{b}}~\mathbb{E}_{z, b}[\log (1-D_{aux}(G_b(z, b)))]
\vspace{-3pt}
\end{align*}

This loss updates $G_b$ so that $D_{aux}$ assigns a high background probability to the generated background patches.

\vspace{-1pt}
\subsubsection{Parent stage}\label{sec:pstage}

As explained previously, we model the real data distribution $p_{data}(x)$ through a two-level foreground generation process via the parent and child stages. The parent stage can be viewed as modeling higher-level information about an object like its shape, and the child stage, \emph{conditioned on the parent stage}, as modeling lower-level information like its color/texture.

Capturing multi-level information in this way can have potential advantages. First, it makes the overall image generation process more principled and easier to interpret; different sub-networks of the model can focus only on synthesizing entities they are concerned with, in contrast to the case where the entire network performs single-shot image generation. Second, for fine-grained generation, it should be easier for the model to generate appearance details \emph{conditioned} on the object's shape, without having to worry about the background and other variations.  With the same reasoning, such hierarchical features---parent capturing shape and child capturing appearance---should also be beneficial for fine-grained categorization compared to a flat-level feature representation.

We now discuss the working details of the parent stage. As shown in Fig.~\ref{fig:main}, $G_p$, which consists of a series of convolutional layers and residual blocks, maps $z$ and $p$ to feature representation $F_p$. As discussed previously, the requirement from this stage is only to generate a foreground entity, and stitch it to the existing background $\mathcal{B}$. Consequently, two generators $G_{p,f}$ and $G_{p,m}$ transform $F_p$ into parent foreground ($\mathcal{P}_f$) and parent mask ($\mathcal{P}_m$) respectively, so that $P_m$ can be used to stitch $P_f$ on $B$, to obtain the parent image $\mathcal{P}$:
$$\mathcal{P} = \mathcal{P}_{f,m} + \mathcal{B}_{m}$$
where $\mathcal{P}_{f,m} = \mathcal{P}_m \odot \mathcal{P}_f$ and
$\mathcal{B}_{m} = (1 - \mathcal{P}_m) \odot \mathcal{B}$ denote masked foreground and inverse masked background image respectively; see green arrows in Fig.~\ref{fig:main}. This idea of generating a mask and using it for stitching is inspired by LR-GAN~\cite{yang-iclr17}.

We again employ a discriminator at the parent stage, and denote it as $D_p$. Its functioning however, differs from the discriminators employed at the other stages. This is because in contrast to the background and child stages where we know the true distribution to be modeled, the true distribution for $\mathcal{P}$ or $\mathcal{P}_{f,m}$ is unknown (i.e., we have real patch samples of background and real image samples of the object, but we do not have any real intermediate image samples in which the object exhibits one factor like shape but lacks another factor like appearance). Consequently, we cannot use the standard GAN objective to train $D_p$.

Thus, we only use $D_p$ to induce the parent code $p$ to represent the hierarchical concept i.e., the object's shape. With no supervision from image labels, we exploit information theory to discover this concept in a completely unsupervised manner, similar to InfoGAN \cite{chen-nips16}. Specifically, we maximize the mutual information $I(p,\mathcal{P}_{f,m})$, with $D_p$ approximating the posterior $P(p|\mathcal{P}_{f,m})$:
\vspace{-3pt}
\begin{equation*}\label{lp}
 \mathcal{L}_{p} = \mathcal{L}_{p\textunderscore info} = \max\limits_{D_p,G_{p,f},G_{p,m}}~\mathbb{E}_{z,p}[\log D_p(p|\mathcal{P}_{f,m})]
 \vspace{-3pt}
\end{equation*}

We use $\mathcal{P}_{f,m}$ instead of $\mathcal{P}$ so that $D_p$ makes its decision solely based on the foreground object (shape) and not get influenced by the background. In simple words, $D_p$ is asked to reconstruct the latent hierarchical category information ($p$) from $\mathcal{P}_{f,m}$, which already has this information encoded during its synthesis. Given our constraints from Sec.~\ref{rel} that there are less parent categories than child ones ($N_p<N_c$) and multiple child codes share the same parent code, FineGAN tries encoding $p$ into $\mathcal{P}_{f,m}$ as an attribute that: (i) by itself cannot capture all fine-grained category details, and (ii) is \emph{common} to multiple fine-grained categories, which is the essence of hierarchy. 

\subsubsection{Child stage}\label{sec:cstage}

The result of the previous stages is an image that is a composition of the background and object's outline.  The task that remains is filling in the outline with appropriate texture/color to generate the final fine-grained object image.

As shown in Fig.~\ref{fig:main}, we encode the color/texture information about the object with child code $c$, which is itself conditioned on parent code $p$. Concatenated with $F_p$, the resulting feature chunk is fed into $G_c$, which again consists of a series of convolutional and residual blocks. Analogous to the parent stage, two generators $G_{c,f}$ and $G_{c,m}$ map the resulting feature representation $F_c$ into child foreground ($\mathcal{C}_{f}$) and child mask ($\mathcal{C}_{m}$) respectively. The stitching process to obtain the complete child image $\mathcal{C}$ is:
\vspace{-4pt}
\begin{equation*}
    \mathcal{C} = \mathcal{C}_{f,m} + \mathcal{P}_{c,m}
    \vspace{-3pt}
\end{equation*}
where $\mathcal{C}_{f,m} = \mathcal{C}_m \odot \mathcal{C}_f$, and $\mathcal{P}_{c,m} = (1 - \mathcal{C}_m) \odot \mathcal{P}$.

We now discuss the requirements for the child stage discriminative networks, $D_{adv}$ and $D_c$: (i) discriminate between real samples from $\mathcal{X}$ and fake samples from the generative distribution using $D_{adv}$; (ii) use $D_c$ to approximate the posterior $P(c|C_{f,m})$ to associate the latent code $c$ to a fine-grained object detail like color and texture. The loss function can hence be broken down into two components $\mathcal{L}_{c} = \mathcal{L}_{c\textunderscore adv} + \mathcal{L}_{c\textunderscore info}$, where:
\vspace{-2pt}
\begin{equation*}
\resizebox{1.05 \columnwidth}{!} {\resizebox{\hsize}{!}{
    \hspace{-2pt}$\mathcal{L}_{c\textunderscore adv} = \min\limits_{G_c}\max\limits_{D_{adv}}~\mathbb{E}_{x}[\log (D_{adv}(x))] +  \mathbb{E}_{z,b,p,c}[\log (1 - D_{adv}(\mathcal{C}))],$
}}
\end{equation*}
\vspace{-0.5cm}
\begin{equation*}
\resizebox{1.05 \columnwidth}{!} {\resizebox{\hsize}{!}{
    \hspace{-2pt}$\mathcal{L}_{c\textunderscore info} = \max\limits_{D_c,G_{c,f},G_{c,m}}~\mathbb{E}_{z,p,c}[\log D_c(c|\mathcal{C}_{f,m})].~~~~~~~~~~~~~~~~~~~~~~~~~~~~~~~~~~~~~$
}}\vspace{-1pt}
\label{lc}
\end{equation*}

Again, we use $\mathcal{C}_{f,m}$ instead of $\mathcal{C}$ so that $D_c$ makes its decision solely based on the object (color/texture and shape) and not get influenced by the background.  With shape already captured though the parent code $p$, the child code $c$ can now solely focus to correspond to the texture/color inside the shape.

\subsection{Fine-grained object category discovery}

Given our trained FineGAN model, we can now use it to compute features for the real images $x_i \in \mathcal{X}$ to cluster them into fine-grained object categories.  In particular, we can make use of the final synthetic images $\{\mathcal{C}_j\}$ and their associated parent and child codes to learn a mapping from images to codes.  Note that we cannot directly use the parent and child discriminators $D_p$ and $D_c$---which each categorize $\{\mathcal{P}_{f,m}\}$ and $\{\mathcal{C}_{f,m}\}$ into one of the parent and child codes respectively----on the real images due to the unavailability of real foreground masks. Instead, we train a pair of convolutional networks ($\phi_p$ and $\phi_c$) to predict the parent and child codes of the final set of synthetic images $\{\mathcal{C}_j\}$:

\begin{enumerate}[noitemsep]
 \item Randomly sample a batch of codes: $z \sim \mathcal{N} (0,1)$, $p \sim p_p$, $c \sim p_c$, $b \sim p_b$ to generate child images $\{\mathcal{C}_j\}$.
 \item Feed forward this batch through $\phi_p$ and $\phi_c$. Compute cross-entropy loss $CE(p, \phi_p(\mathcal{C}_j))$ and $CE(c, \phi_c(\mathcal{C}_j))$.

 \item Update $\phi_p$ and $\phi_c$. Repeat till convergence.
\vspace{-3pt}
\end{enumerate}

To accurately predict parent code $p$ from $\mathcal{C}_j$, $\phi_p$ has to solely focus on the object's shape as no sensible supervision can come from the randomly chosen background and child codes. With similar reasoning, $\phi_c$ has to solely focus on the object's appearance to accurately predict child code $c$. Once $\phi_p$ and $\phi_c$ are trained, we use them to extract features for each real image $x_i \in \mathcal{X}$. Finally, we use their concatenated features to group the images with $k$-means clustering.

\section{Experiments}\label{exp}

We first evaluate FineGAN's ability to disentangle and generate images of fine-grained object categories. We then evaluate FineGAN's learned features for fine-grained object clustering with real images.

\begin{figure*}[t!]
    \centering
    \includegraphics[width=0.98\textwidth]{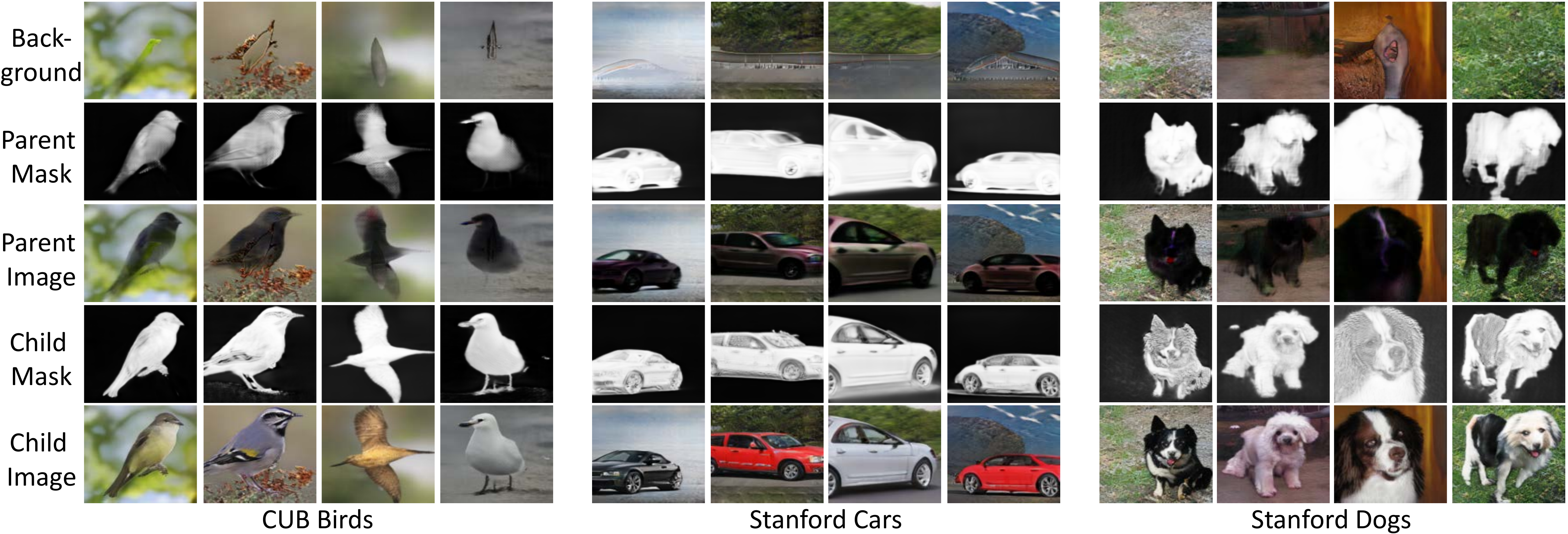}
    \caption{\textbf{FineGAN's stagewise image generation.} Background stage generates a background which is retained over the child and parent stages. Parent stage generates a hollow image with only the object's shape, and child stage fills in the appearance to complete the image.}
    \label{fig:all_stage}
    \vspace{-0.1in}
\end{figure*}

\vspace{-10pt}
\paragraph{Datasets and implementation details.} We evaluate on three fine-grained datasets: (1) \textbf{CUB}~\cite{wah-tech11}: 200 bird classes. We use the entire dataset (11,788 images); (2) \textbf{Stanford Dogs}~\cite{khosla-FGVC11}: 120 dog classes. We use its train data (12,000 images); (3) \textbf{Stanford Cars}~\cite{krause-DRR2013}: 196 car classes. We use its train data (8,144 images).  \emph{We do not use any of the provided labels for training.  The labels are only used for evaluation.}
Number of parents and children are set as: (1) CUB: $N_p$ = 20, $N_c$ = 200; (2) Stanford dogs: $N_p$ = 12, $N_c$ = 120; and (3) Cars: $N_p$ = 20, $N_c$ = 196.
$N_c$ matches the ground-truth number of fine-grained classes per dataset. We set $\lambda$ = 10, $\beta$ = 1 and $\gamma$ = 1 for all datasets.

\subsection{Fine-grained image generation}

We first analyze FineGAN's image generation in terms of realism and diversity.  We compare to:
\begin{itemize}[noitemsep,leftmargin=*]
    \item \textbf{Simple-GAN}: Generates a final image ($\mathcal{C}$) in one shot without the parent and background stages. Only has $\mathcal{L}_{c\textunderscore adv}$ loss at the child stage. This baseline helps gauge the importance of disentanglement learned by $\mathcal{L}_{c\textunderscore info}$. For fair comparison, we use FineGAN's backbone architecture.
    \item \textbf{InfoGAN}~\cite{chen-nips16}: Same as Simple-GAN but with additional $\mathcal{L}_{c\textunderscore info}$. This baseline helps analyze the importance of hierarchical disentanglement between background, shape, and appearance during image generation, which is lacking in InfoGAN. $N_c$ is set to be the same as FineGAN for each dataset. We again use FineGAN's backbone architecture. 
    \item \textbf{LR-GAN}~\cite{yang-iclr17}: It also generates an image stagewise, which is similar to our approach. But its stages only consist of foreground and background, and that too controlled by single random vector $z$.
    \item \textbf{StackGAN-v2}~\cite{stackgan2}: Its unconditional version generates images at multiple scales with $\mathcal{L}_{c\textunderscore adv}$ at each scale. This helps gauge how FineGAN fares against a state-of-the-art unconditional image generation approach.
\end{itemize}

For LR-GAN and StackGAN-v2, we use the authors' publicly-available code.  We evaluate image generation using Inception Score (\textbf{IS})~\cite{salimans-nips16} and Frechet Inception Distance (\textbf{FID})~\cite{fid}, which are computed on 30K randomly generated images (equal number of images for each child code $c$), using an Inception Network fine-tuned on the respective datasets~\cite{barratt-arxiv2018}. We evaluate on $128$ x $128$ generated images for all methods except LR-GAN, for which $64$ x $64$ generated images give better performance.

\begin{figure*}[t!]
    \centering
    \includegraphics[width=0.97\textwidth]{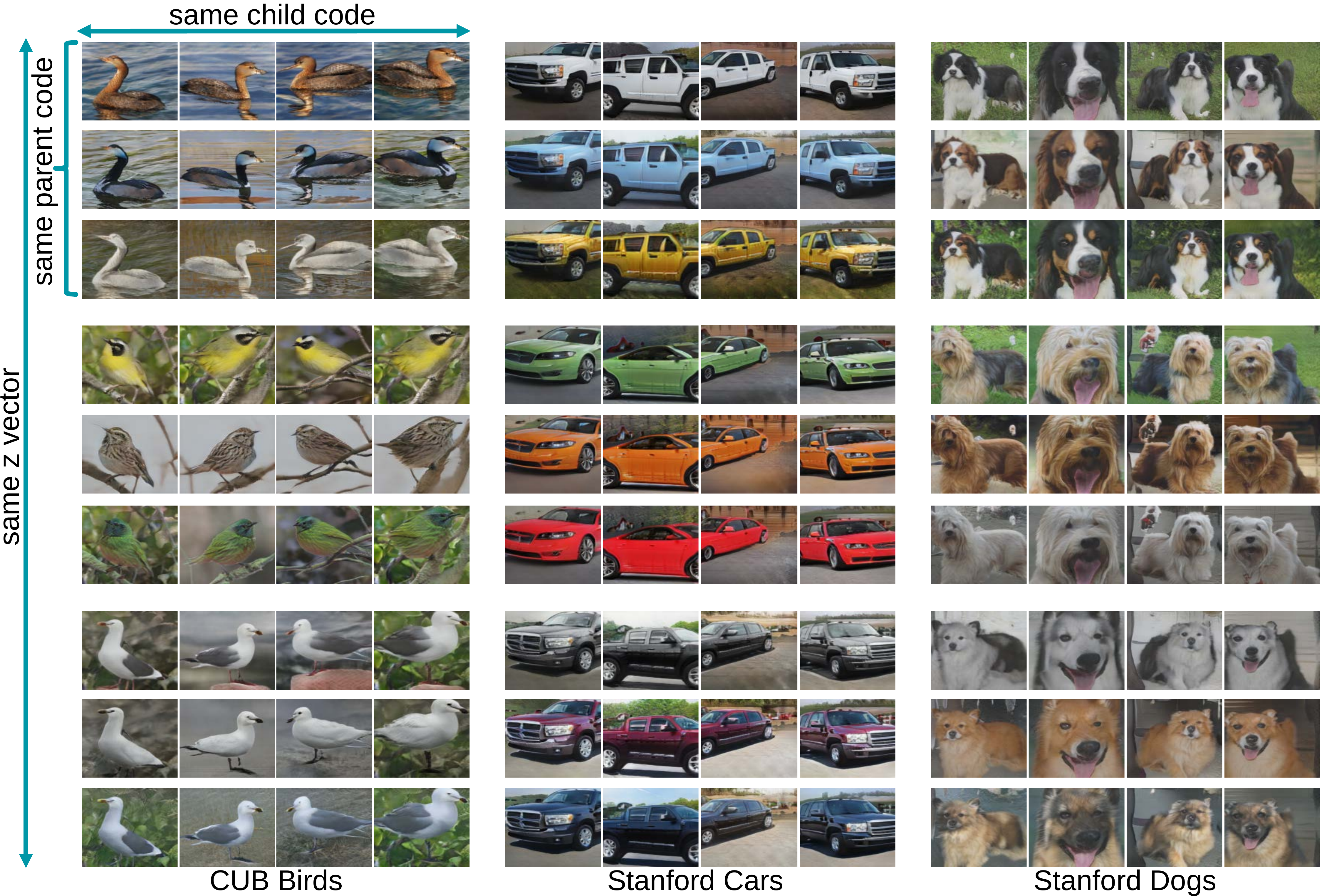}
    \caption{\textbf{Varying $p$ vs.~$c$ vs.~$z$.} Every three rows correspond to the same parent code $p$ and each row has a different child code $c$. For the same parent, the object's shape remains consistent while the appearance changes with different child codes. For the same child, the appearance remains consistent. Each column has the same random vector $z$ -- we see that it controls the object's pose and position.}
    \label{fig:zc_vis}
    \vspace{-0.1in}
\end{figure*}

\subsubsection{Quantitative evaluation of image generation}\label{sec:quant}

FineGAN obtains Inception Scores and FIDs that are favorable when compared to the baselines (see Table~\ref{table:inception_score}), which shows it can generate images that closely match the real data distribution. 

In particular, the lower scores by Simple-GAN, LR-GAN, and StackGAN-v2 show that relying on a single adversarial loss can be insufficient to model fine-grained details. Both FineGAN and InfoGAN learn to associate a $c$ code to a variation factor ($\mathcal{L}_{c\textunderscore info}$) to generate more detailed images.  But by further disentangling the background and object shape (parent), FineGAN learns to generate more diverse images. LR-GAN also generates an image stagewise but we believe it has lower performance as it only separates foreground and background, which appears to be insufficient for capturing fine-grained details. These results strongly suggest that FineGAN's hierarchical disentanglement is important for better fine-grained image generation.

\begin{table}[t!]
    \begin{center}
        \hspace*{-5pt}
		\tabcolsep=0.1cm
		\scriptsize
		\resizebox{0.49\textwidth}{!}{
		\begin{tabular}{ c | c  c  c | c  c  c }
		    & \multicolumn{3}{c}{IS} & \multicolumn{3}{|c}{FID}\\
			\hline
			 & Birds & Dogs & Cars & Birds & Dogs & Cars\\
			\hline
			Simple-GAN & 31.85 $\pm$ 0.17 & 6.75 $\pm$ 0.07 & 20.92 $\pm$ 0.14 & 16.69   & 261.85 & 33.35 \\
			InfoGAN~\cite{chen-nips16} & 47.32 $\pm$ 0.77   & 43.16 $\pm$ 0.42 & 28.62 $\pm$ 0.44 & 13.20  & 29.34 & 17.63\\
			LR-GAN~\cite{yang-iclr17} & 13.50 $\pm$ 0.20   & 10.22 $\pm$ 0.21 & 5.25 $\pm$ 0.05 & 34.91  & 54.91 & 88.80\\
			StackGANv2~\cite{stackgan2} & 43.47 $\pm$ 0.74   & 37.29 $\pm$ 0.56 & \textbf{33.69 $\pm$ 0.44} & 13.60   & 31.39 & 16.28\\
			FineGAN (ours) & \textbf{52.53 $\pm$ 0.45}   & \textbf{46.92 $\pm$ 0.61} & 32.62 $\pm$ 0.37 & \textbf{11.25}   & \textbf{25.66} & \textbf{16.03}\\
			\hline
		\end{tabular}}
		\caption{Inception Score (higher is better) and FID (lower is better). FineGAN consistently generates diverse and real images that compare favorably to those of state-of-the-art baselines.}
		\label{table:inception_score}
		\vspace{-0.2in}
	\end{center}
\end{table}

\begin{table}[t!]
	\begin{center}
	    \tabcolsep=0.1cm
		\footnotesize
		\begin{tabular}{ c | c  c  c  c  c}
			& $N_p$=20 & $N_p$=10  & $N_p$=40 & $N_p$=5 & $N_p$=mixed  \\	
			\hline
			Inception Score (CUB) & \textbf{52.53} & 52.11  & 49.62 & 46.68 & 51.83  \\
			\hline
		\end{tabular}
		\caption{Varying number of parent codes $N_p$, with number of children $N_c$ fixed to 200. FineGAN is robust to a wide range of $N_p$.}
		\label{table:p_comp}
		\vspace{-0.2in}
	\end{center}
\end{table}

\vspace{-10pt}
\paragraph{How sensitive is FineGAN to the number of parents?} Table~\ref{table:p_comp} shows the Inception Score (IS) on CUB of FineGAN trained with varying number of parents while keeping the number of children fixed (200).  IS remains consistently high unless we have very small ($N_p$=5) or large ($N_p$=40) number of parents. With very small $N_p$, we limit diversity in the number of object shapes, and with very high $N_p$, the model has less opportunity to take advantage of the implicit hierarchy in the data. With variable number of children per parent ($N_p$=mixed: 6 parents with 5 children, 3 parents with 20 children, and 11 parents with 10 children), IS remains high, which shows there is no need to have the same number of children for each parent. These results show that FineGAN is robust to a wide range of parent choices.

\subsubsection{Qualitative evaluation of image generation}\label{qual_eval}

We next qualitatively analyze FineGAN's (i) image generation process; (ii) disentanglement of the factors of variation; and provide (iii) in-depth comparison to InfoGAN.

\vspace{-10pt}
\paragraph{Image generation process.}
Fig.~\ref{fig:all_stage} shows the intermediate images generated for CUB, Stanford Cars, and Stanford Dogs. The background images (1st row) capture the context of each dataset well; e.g., they contain roads for cars, gardens or indoor scene for dogs, leafy backgrounds for birds. The parent stage produces parent masks that capture each object's shape (2nd row), and a textureless, hollow entity as the parent image (3rd row) together with the background. The final child stage produces a more detailed mask (4th row) and the final composed image (last row), which has the same foreground shape as that of the parent image with added texture/color details.  Note that the generation of accurate masks at each stage is important for the final composed image to retain the background, and is obtained without any mask supervision during training.  We present additional quantitative analyses on the quality of the masks in the supplementary material.

\vspace{-10pt}
\paragraph{Disentanglement of factors of variation.}
Fig.~\ref{fig:zc_vis} shows the discovered grouping of parent and child codes by FineGAN. Each row corresponds to different instances with the same child code. Two observations can be made as we move left to right: (i) there is a consistency in the appearance and shape of the foreground objects; (ii) background changes slightly, giving an impression that the background across a row belongs to the same class, but with slight modifications.  For each dataset, each set of three rows corresponds to three distinct children of the same parent, which is evident from their common shape. Notice that different child codes for the same parent can capture fine-grained differences in the appearance of the foreground object (e.g., dogs in the third row differ from those in first only because of small brown patches; similarly, birds in the 7th and last rows differ only in their wing color).  Finally, the consistency in object viewpoint and pose along each column shows that FineGAN has learned to associate $z$ with these factors.

\begin{figure}[t!]
    \centering
    \includegraphics[width=0.48\textwidth]{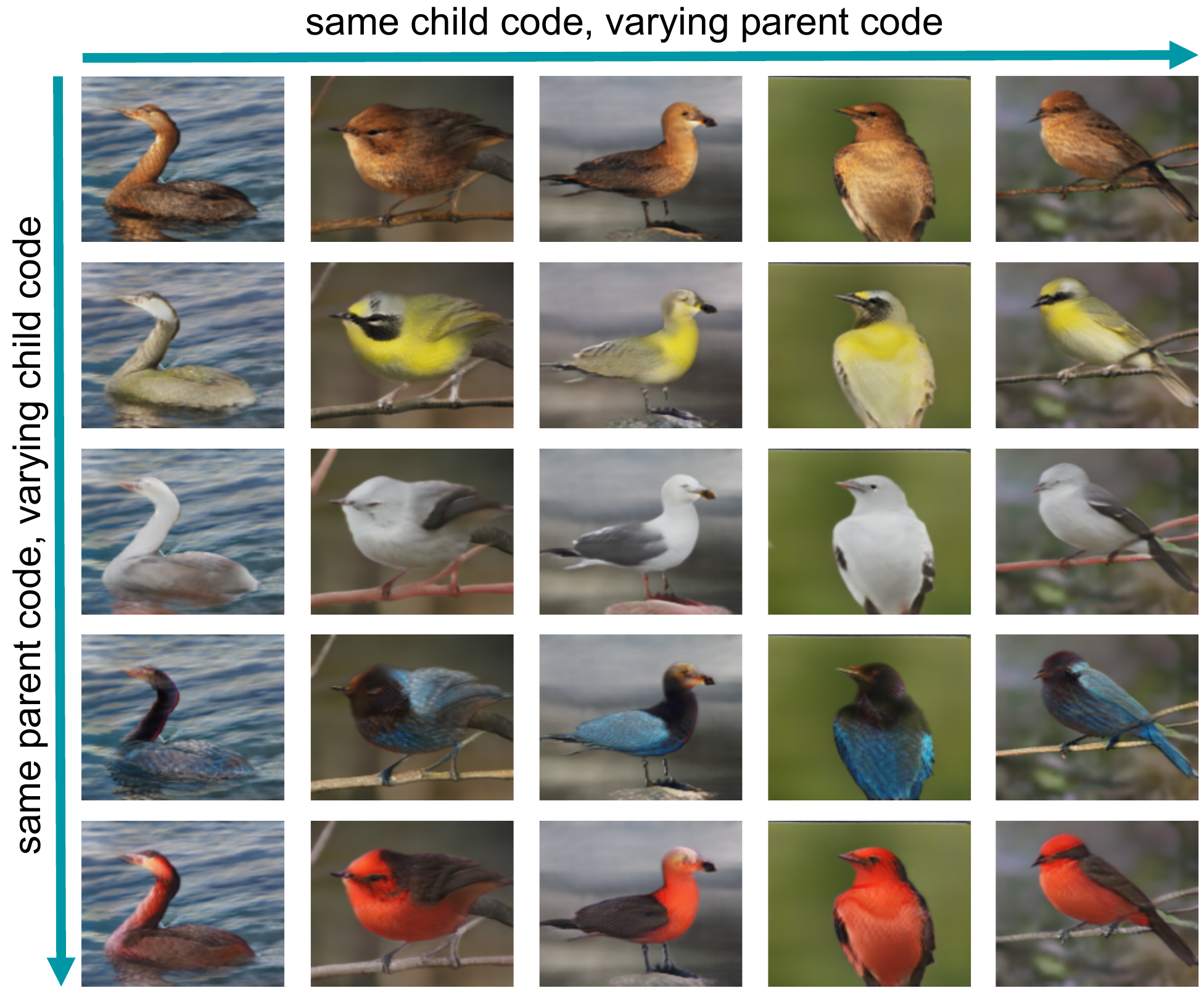}
    \caption{\textbf{Disentanglement of parent vs.~child codes.} Shape is retained over the column, appearance is retained over the row.}
    \label{pvsc}
    \vspace{-0.1in}
\end{figure}

\vspace{-10pt}
\paragraph{Disentanglement of parent vs.~child.}
Fig.~\ref{pvsc} further analyzes the disentanglement of parent (shape) and child code (appearance). Across the rows, we vary parent code $p$ while keeping child code $c$ constant, which changes the bird's shape but keeps the texture/color the same. Across the columns, we vary child code $c$ while keeping parent code $p$ constant, which changes the bird's color/texture but keeps the shape the same. This result illustrates the control that FineGAN has learned \emph{without any corresponding supervision} over the shape and appearance of a bird. Note that we keep background code $b$ to be same across each column.

\vspace{-10pt}
\paragraph{Disentanglement of background vs.~foreground.}
The figure below shows disentanglement of background from object.  In (a), we keep background code $b$ constant and vary the parent and child code, which generates different birds over the same background.  In (b), we keep the parent and child codes constant and vary the background code, which generates an identical bird with different backgrounds.
\vspace{-10pt}
\begin{figure}[h!]
    \centering
    \includegraphics[width=0.48\textwidth]{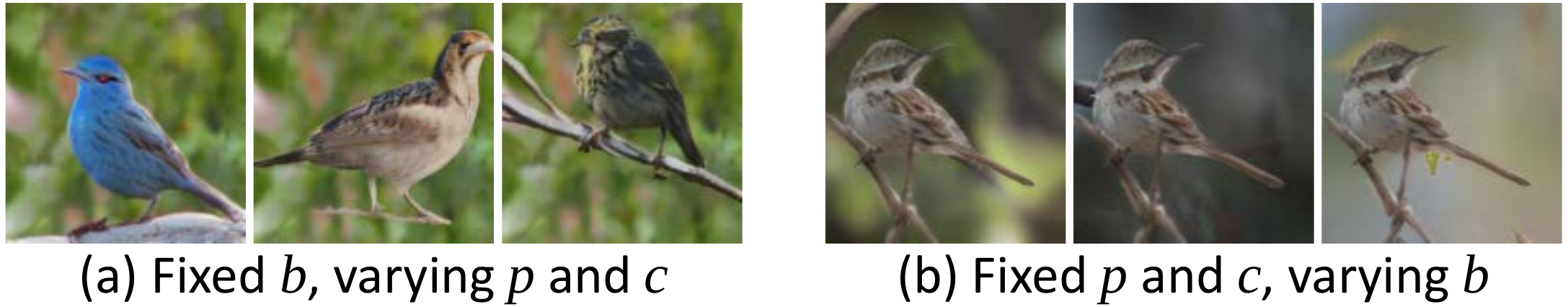}
\end{figure}

\begin{figure}[t!]
    \centering
    \includegraphics[width=0.48\textwidth]{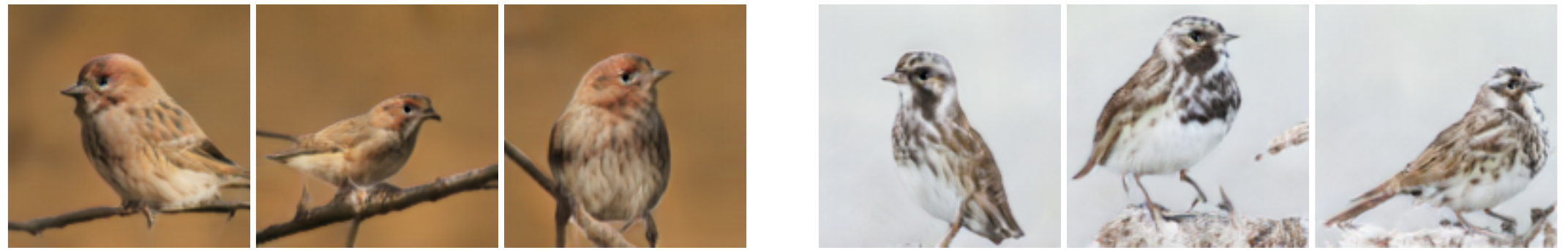}
    \caption{\textbf{InfoGAN results.} Images in each group have same child code.  The birds are the same, but so are their backgrounds. This strongly suggests InfoGAN takes background into consideration when categorizing the images. In contrast, FineGAN's generated images (Fig.~\ref{fig:zc_vis}) for same $c$ show reasonable variety in background.}
    \label{infogan}
    \vspace{-0.1in}
\end{figure}

\vspace{-15pt}
\paragraph{Comparison with InfoGAN.} In InfoGAN~\cite{chen-nips16}, the latent code prediction is based on the complete image, in contrast to FineGAN which uses the masked foreground. Due to this, InfoGAN's child code prediction can be biased by the background (see Fig.~\ref{infogan}). Furthermore, InfoGAN~\cite{chen-nips16} does not hierarchically disentangle the latent factors.  To enable InfoGAN to model the hierarchy in the data, we tried conditioning its generator on both the parent and child codes, and ask the discriminator to predict both. This improves performance slightly (IS: 48.06, FID: 12.84 for birds), but is still worse than FineGAN.  This shows that simply adding a parent code constraint to InfoGAN does not lead it to produce the hierarchical disentanglement that FineGAN achieves.

\subsection{Fine-grained object category discovery}

We next evaluate FineGAN's learned features for clustering real images into fine-grained object categories.  We compare against the state-of-the-art deep clustering approaches \textbf{JULE}~\cite{yang-cvpr16} and  \textbf{DEPICT}~\cite{dizaji-iccv17}. To make them even more competitive, we also create a JULE variant with ResNet-50 backbone (\textbf{JULE-ResNet-50}) and DEPICT with double the number of filters in each conv layer (\textbf{DEPICT-Large}). We use code provided by the authors. All methods cluster the same image regions.

\begin{table}[t!]
	\begin{center}
	    \tabcolsep=0.15cm
		\scriptsize
		\begin{tabular}{ c | c  c  c | c  c  c }
		& \multicolumn{3}{c}{NMI} & \multicolumn{3}{|c}{Accuracy} \\
			\hline    	
			 & Birds  & Dogs & Cars & Birds  & Dogs & Cars \\	
			\hline
			JULE~\cite{yang-cvpr16}   & 0.204 & 0.142 & 0.232 & 0.045   & 0.043 & 0.046 \\
			JULE-ResNet-50~\cite{yang-cvpr16}  & 0.203 & 0.148 & 0.237 & 0.044  & 0.044 & 0.050\\
			DEPICT~\cite{dizaji-iccv17}   & 0.290 & 0.182 & 0.329 & 0.061  & 0.052 & 0.063\\
			DEPICT-Large~\cite{dizaji-iccv17}   & 0.297 & 0.183 & 0.330 & 0.061  & 0.054 & 0.062\\
			Ours & \textbf{0.403}  & \textbf{0.233} & \textbf{0.354} & \textbf{0.126}  & \textbf{0.079} & \textbf{0.078}\\
			\hline
		\end{tabular}
		\caption{Our approach outperforms existing clustering methods.}
		\label{table:nmi}
		\vspace{-0.2in}
	\end{center}
\end{table}

For evaluation we use Normalized Mutual Information (\textbf{NMI})~\cite{xu-sigir03} and \textbf{Accuracy} (of best mapping between cluster assignments and true labels) following~\cite{dizaji-iccv17}.  Our approach outperforms the baselines on all three datasets (see Table~\ref{table:nmi}).  This indicates that FineGAN's features learned for hierarchical image generation are better able to capture the fine-grained object details necessary for fine-grained object clustering.  JULE and DEPICT are unable to capture those details to the same extent; instead, they focus more on high-level details like background and rough shape (see supp. for examples). Increasing their capacity (JULE-RESNET-50 and DEPICT-Large) gives little improvement. Finally, if we only use our child code features, then performance drops ($0.017$ in Accuracy on birds). This shows that the parent code and child code features are complementary and capture different aspects (shape vs.~appearance).

\section{Discussion and limitations}
\vspace{-1pt}

There are some limitations of FineGAN worth discussing.  First, although we have shown that our model is robust to a wide range of number of parents (Table~\ref{table:p_comp}), it along with the number of children are hyperparameters that a user must set, which can be difficult when the true number of categories is unknown (a problem common to most unsupervised grouping methods).  Second, the latent modes of variation that FineGAN discovers may not necessarily correspond to those defined/annotated by a human.  For example, our results in Fig.~\ref{fig:zc_vis} for cars show that the children are grouped based on color rather than car model type.  This highlights the importance of a good evaluation metric for unsupervised methods.  Third, in our current implementation, FineGAN requires bounding boxes to model the background. In preliminary experiments, we observe that approximating the background with patches lying along the border of real images also gives reasonable results. 
Finally, while we significantly outperform unsupervised clustering methods, we are far behind fully-supervised fine-grained recognition methods.  Nonetheless, we feel that this paper has taken important initial steps in tackling the challenging problem of unsupervised fine-grained object modeling.

\vspace{-12pt}
\paragraph{Acknowledgments.} {This work was supported in part by NSF IIS-1751206, IIS-1748387, AWS ML Research Award, Google Cloud Platform research credits, and GPUs donated by NVIDIA.}

{\small
\bibliographystyle{ieee_fullname}
\bibliography{egbib}
}
\newpage
\setcounter{section}{0}

\section*{Appendix}

Here, we provide qualitative clustering results, additional quantitative analysis on the quality of the generated masks, and architecture and training details.  We also provide additional clustering and image generation results.


\begin{figure}[t!]
    \centering
    \includegraphics[width=0.48\textwidth]{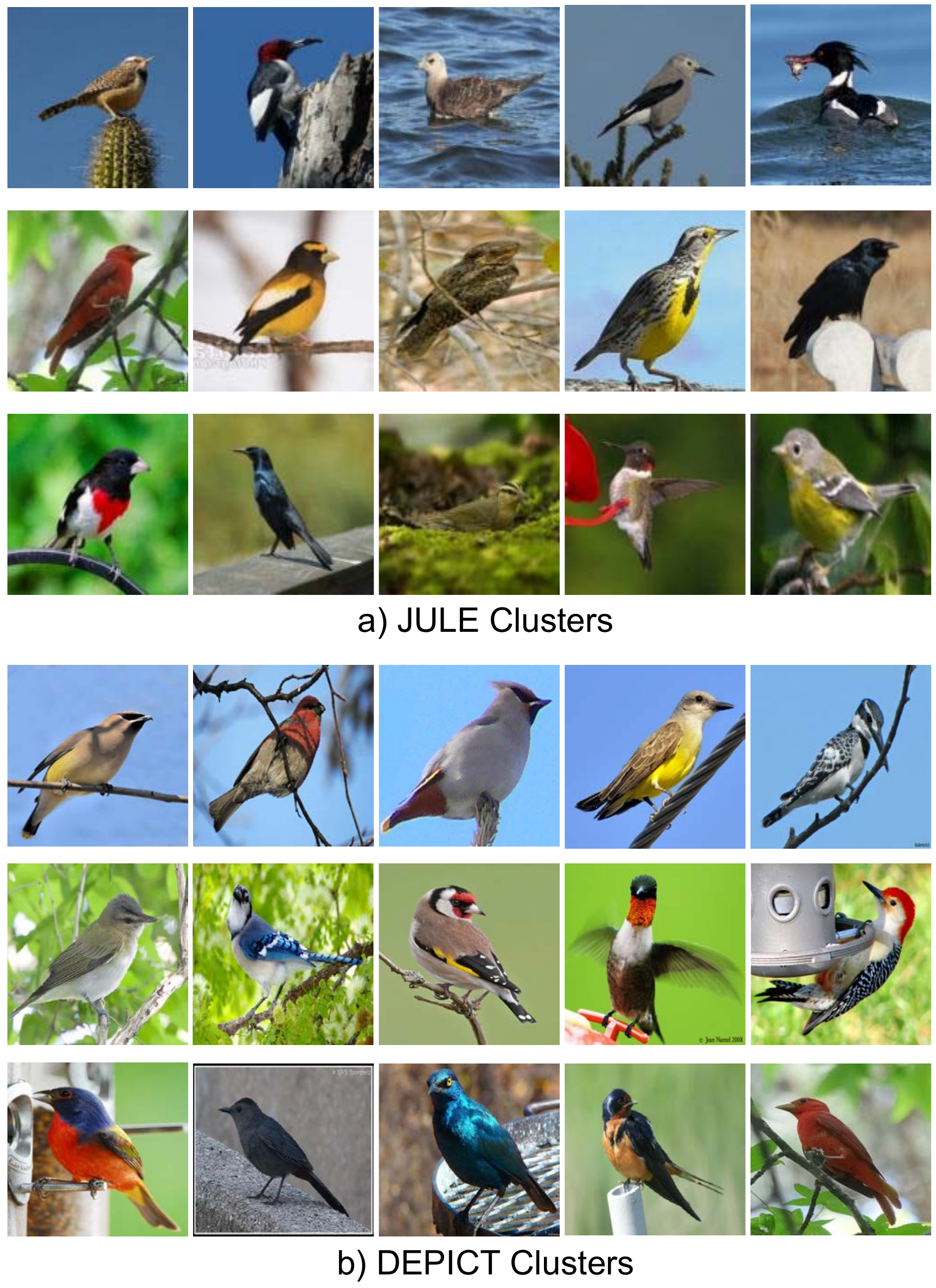}
    \caption{\textbf{JULE and DEPICT clusters.} Each row corresponds to a cluster. The clusters are mainly formed on the basis of non fine-grained elements like background and rough shape.}
    \label{jule_depict_cluster}
    \vspace{-0.1in}
\end{figure}

\begin{figure}[t!]
    \centering
    \includegraphics[width=0.48\textwidth]{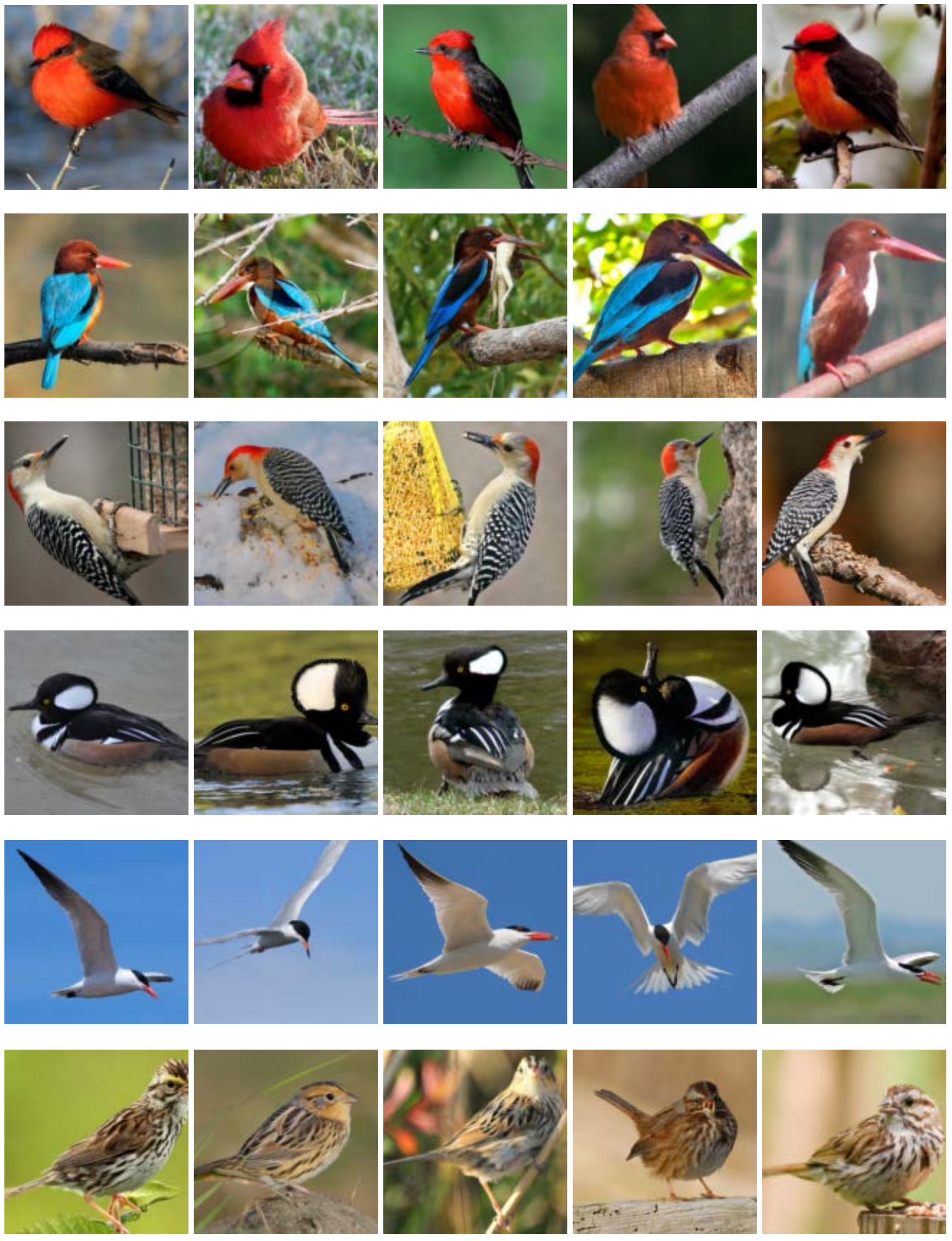}
    \caption{\textbf{Our clusters.} Each row corresponds to a cluster. We can see that each cluster captures fine-grained appearance details. Also, the background varies for each cluster which shows that the clustering does not significantly dependent on background.}
    \label{our_cluster}
\end{figure}

\section{Qualitative clustering results}

In this section, we visualize the clusters formed using FineGAN's learned features, JULE~\cite{yang-cvpr16}, and DEPICT~\cite{dizaji-iccv17}. JULE performs alternating deep representation learning and clustering whereas DEPICT minimizes a relative entropy loss with a regularization term for balanced clustering.

In Fig.~\ref{jule_depict_cluster}, each row corresponds to members of a cluster formed by JULE and DEPICT. We can see that these clusters are mainly formed on the basis of background and object shape rather than fine-grained object similarity. For example, the clusters formed by JULE in the first and third rows have consistent blue and green background, respectively. Similarly, the clusters in the second row of JULE and first and third rows of DEPICT have consistent object shape and pose. Overall, these qualitative results reflect the worse quantitative performance of JULE and DEPICT shown in Table 3 of the main paper.

\begin{figure}[t!]
    \centering
    \includegraphics[width=0.48\textwidth]{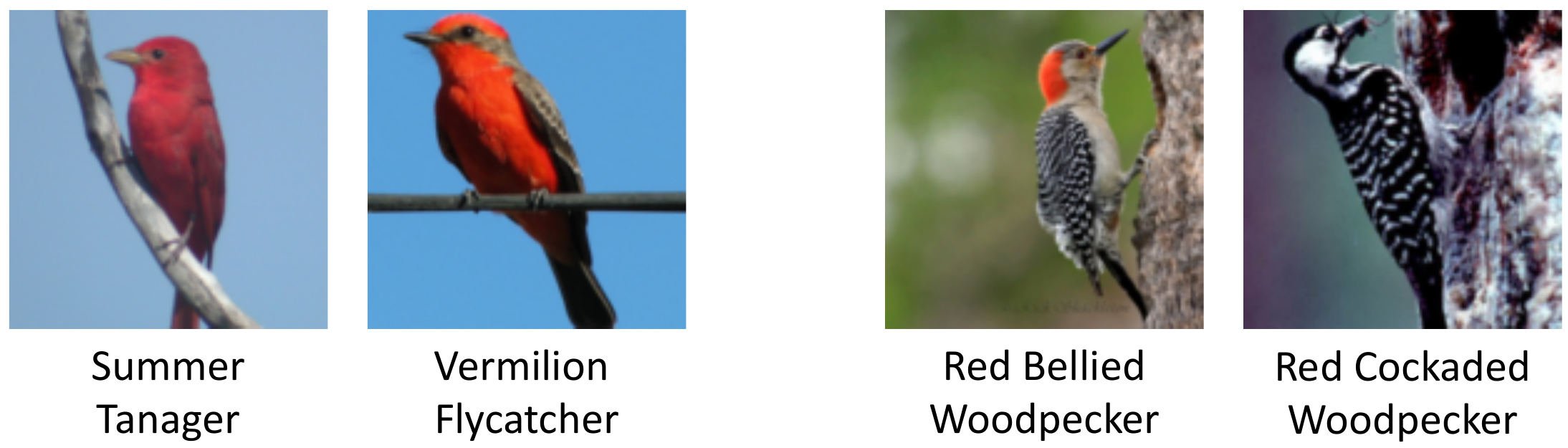}
    \caption{\textbf{Fine-grained clusters.} Our approach is able to group classes which only vary at the fine-grained level into different clusters. For example, `Summer Tanger' and `Vermillion Flycatcher' which only differ by a black stripe on their body are grouped into different clusters. Similarly, `Red Bellied Woodpecker' and `Red Cockaded Woodpecker' are grouped into different clusters.}
    \label{fine_grained_cluster}
    \vspace{-0.1in}
\end{figure}

Next, in Fig.~\ref{our_cluster}, we show our clusters formed using FineGAN's learned features (obtained by concatenating the penultimate layer outputs of $\phi_p$ and $\phi_c$); again, each row corresponds to members of a cluster. Here, many clusters have a representative shape and appearance, which correspond to a true fine-grained category. For example, the cluster in the first row has birds which have consistent shape with a red body and black stripe. Also, the background varies within each cluster which indicates that background is not used as a significant cue for clustering. These results show that FineGAN's learned features are able to better capture fine-grained details. To further investigate, we analyzed specific classes in CUB~\cite{wah-tech11} which only differ at the fine-grained level that our approach is able to discover as separate clusters. Example classes are shown in Fig.~\ref{fine_grained_cluster}. Our approach creates clusters in which the majority of images of `Summer Tanager' and `Vermilion Flycatcher' are placed in separate clusters. We observe similar results for the `Red Bellied Woodpecker' and `Red Cockaded Woodpecker' classes.

\section{Quantitative evaluation of mask generation}
In Fig. 3 of the main paper, we show the masks formed by FineGAN at the parent and child stages. In this section, we quantitatively evaluate the quality of the masks.

We first train a network which takes our final generated image $\mathcal{C}$ as input and predicts the corresponding parent mask which we obtained automatically as part of the generation process. We then use this network to predict the foreground mask on real images.  To evaluate, we use the entire CUB dataset~\cite{wah-tech11}, which has ground-truth foreground masks for birds. Our model obtains a high 75.6\% mIU~\cite{long-cvpr15} when compared against the ground-truth segmentation masks. This shows the accuracy of the masks produced by FineGAN. 
Next, we measure how much of the parent mask is retained at the child stage. Ideally, a high percentage of the parent mask should be retained at the child stage, because our hypothesis is that the child stage should focus on generating the object's appearance conditioned on the shape produced in the parent stage. To measure this, we compute the intersection between the binarized parent and child masks, and normalize the result by the child mask area. This value is 0.70; i.e., a large portion of the changes at the child stage is within the parent mask.  This also shows that most of the background is retained at the child stage.

\section{Attempting to model hierarchy with InfoGAN}

\begin{figure}[t!]
 \centering
 \includegraphics[width=0.45\textwidth]{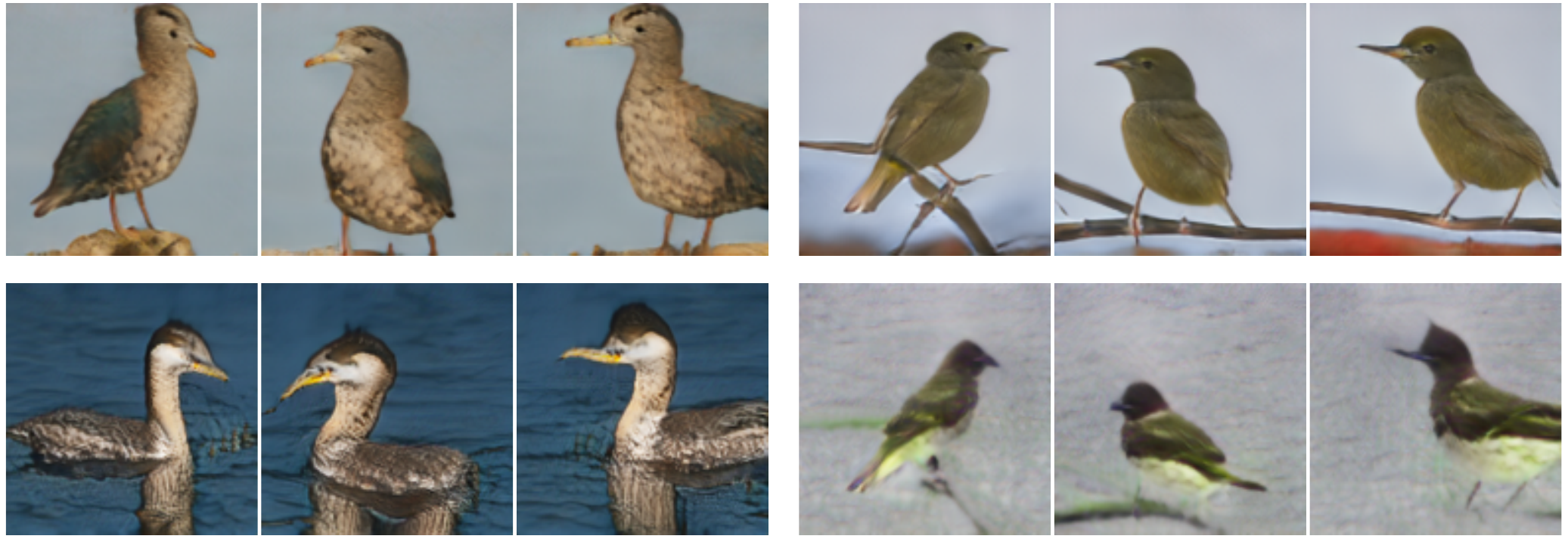}
 \caption{Images generated by an InfoGAN model trained to predict both parent and child codes.  Here all four child groups have the same parent, but do not share a noticeable common pattern (e.g., consistent in either shape, color, or background).}
 \label{pc_infogan}
\end{figure}

As explained in ``Comparison with InfoGAN'' in the main paper, conditioning InfoGAN's generator on both the parent and child codes and asking its discriminator to predict both only results in minor improvement in image generation (IS: 48.06, FID: 12.84 for birds), that is still worse than FineGAN.  Fig.~\ref{pc_infogan} shows qualitative results. Here, all four child groups have the same parent code, but do not seem to share any meaningful attribute (shape, color, or background). This shows that simply adding a parent code constraint to InfoGAN does not lead it to produce the hierarchical disentanglement that FineGAN achieves.

\section{Relaxing relationship constraints during inference}

In Section 3.1 of the main paper, we described the constraints that we impose during training on the relationships between background, parent, and child codes to facilitate the desired entanglement. In this section, we qualitatively reason about the importance of these relationships, and show how these constraints can be relaxed during \emph{inference}.

\begin{figure}[t!]
    \centering
    \includegraphics[width=0.48\textwidth]{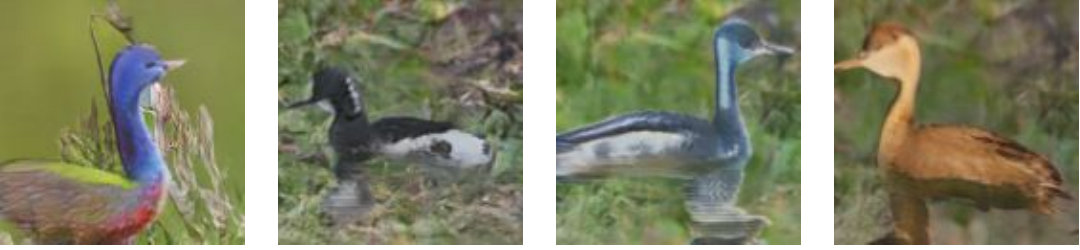}
    \caption{\textbf{Relaxing the latent code relationship constraints during inference.} The generated ducks have unusual green/leafy backgrounds. The first image also has a rare color considering its shape.}
    \label{weird_combo}
    \vspace{-0.1in}
\end{figure}

Fig.~\ref{weird_combo} shows FineGAN's generated images when the background code ($b$) corresponding to a green/leafy background is chosen along with the parent code ($p$) corresponding to a duck-like category. During training, $D_{adv}$ would easily categorize these as fake images since the ducks in the CUB dataset mostly appear in water-like blue backgrounds. Furthermore, the first image in Fig.~\ref{weird_combo} illustrates the need for tying a fixed number of child codes to the same parent code. The combination of blue, green, and red texture is rare for duck-like birds in CUB; and again $D_{adv}$ could use this to categorize the image as fake.  However, during inference, we can relax these constraints as the generator no longer needs to fool the discriminator.

FineGAN can easily generate images like these with `inconsistent' background and texture, only because it has learned a disentangled representation of these factors, and can hence compose an image by choosing any combination of the factors.   Fig.~5 in the main paper shows additional examples (e.g., a red seagull).

\section{Combining factors from multiple real images}
\begin{figure}[t!]
    \centering
    \includegraphics[width=0.48\textwidth]{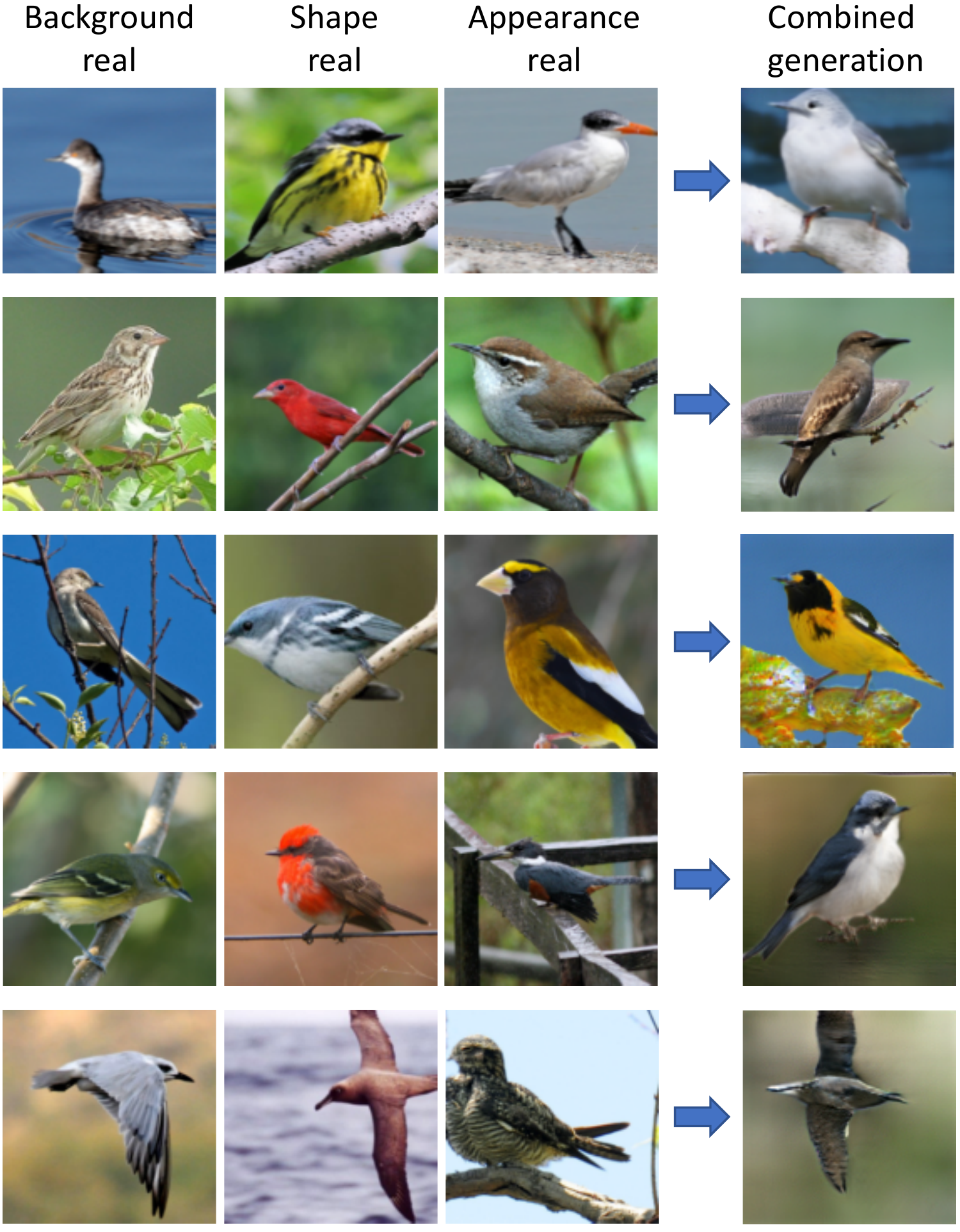}
    \caption{\textbf{Combining factors from multiple real images.} We can generate fake images (fourth column) which combine the background (first column), shape (second column), and appearance (third column) of real images.}
    \label{real_modify}
    \vspace{-0.1in}
\end{figure}

We show additional examples of fusing different factors of variation from multiple real images to synthesize a new image.  Specifically, we use $\phi_c$, $\phi_p$, and similarly-trained $\phi_b$, to compute $c$, $p$, and $b$ codes respectively for different real images.  We then input these codes to FineGAN to generate an image with their combined factors. Fig.~\ref{real_modify} shows images generated in this way.  The 4th column are generated images that have the  background of the 1st column real image, shape of the 2nd column real image, and appearance of the 3rd column real image.  This application of FineGAN could potentially be useful for artists who want to generate new images by combining factors from multiple visual examples.

\section{Additional image generation visualizations}

\begin{figure*}[t!]
    \centering
    \includegraphics[width=0.9\textwidth]{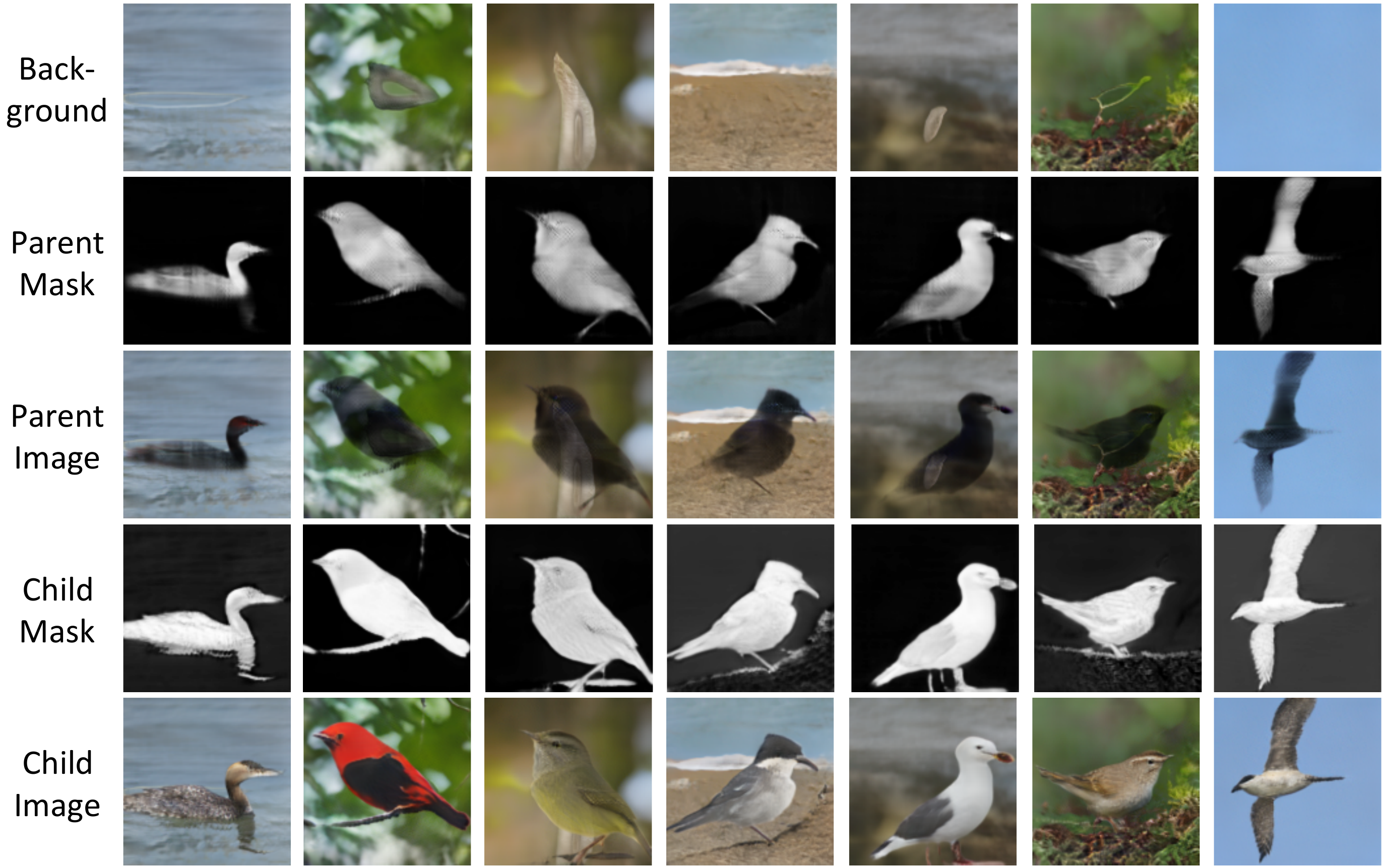}
    \caption{\textbf{FineGAN's stagewise image generation for CUB.} Background stage generates a background which is retained over the child and parent stages. Parent stage generates a hollow image with only the object's shape, and child stage fills in the appearance.}
    \label{fig:all_stage_bird}
    \vspace{0.2in}
\end{figure*}

\begin{figure*}[t!]
    \centering
    \includegraphics[width=0.9\textwidth]{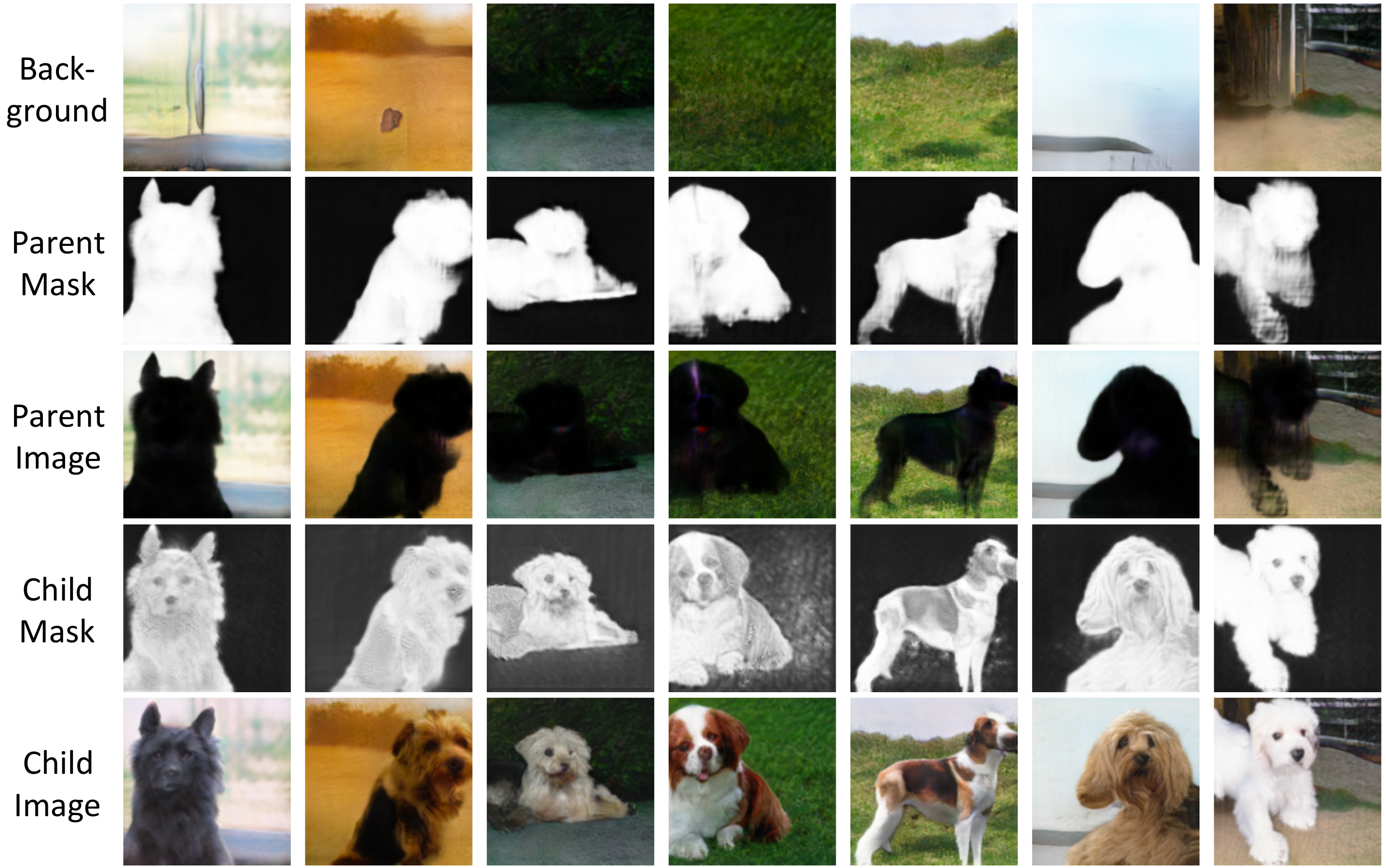}
    \caption{\textbf{FineGAN's stagewise image generation for dogs.} Background stage generates a background which is retained over the child and parent stages. Parent stage generates a hollow image with only the object's shape, and child stage fills in the appearance.}
    \label{fig:all_stage_dog}
\end{figure*}

\begin{figure*}[t!]
    \centering
    \includegraphics[width=1.0\textwidth]{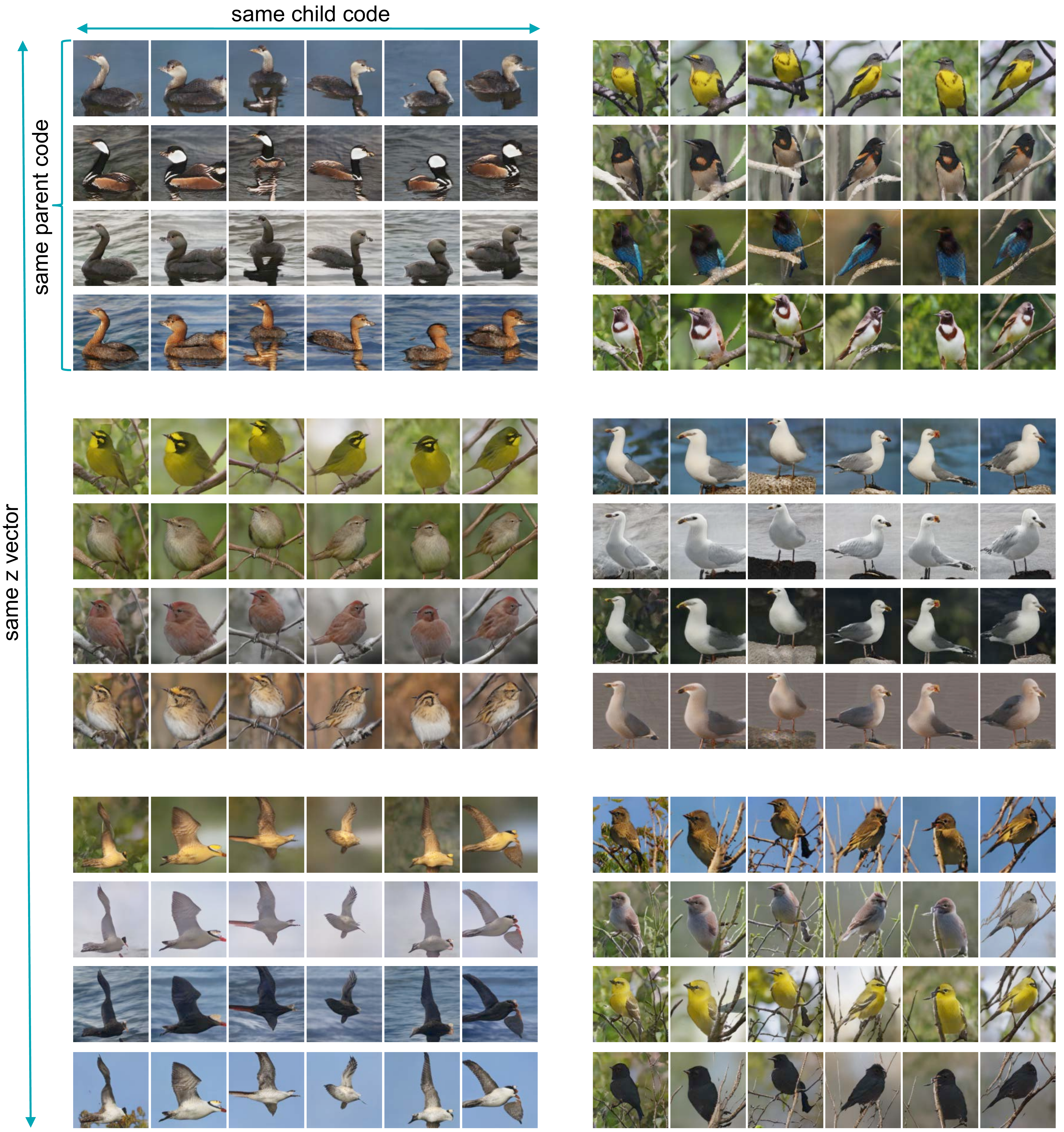}
    \caption{\textbf{Varying $p$ vs.~$c$ vs.~$z$ for CUB.} \emph{We show 6 different parent groups (with different $p$'s), each with 4 children (with different $c$'s).} For the same parent, the object's shape remains consistent while the appearance changes with different child codes. For the same child, the appearance remains consistent. Each column has the same random vector $z$ -- we can see that it controls the object's pose and position. }
    \label{fig:birds_zc_vis}
\end{figure*}

\begin{figure*}[t!]
    \centering
    \includegraphics[width=1.0\textwidth]{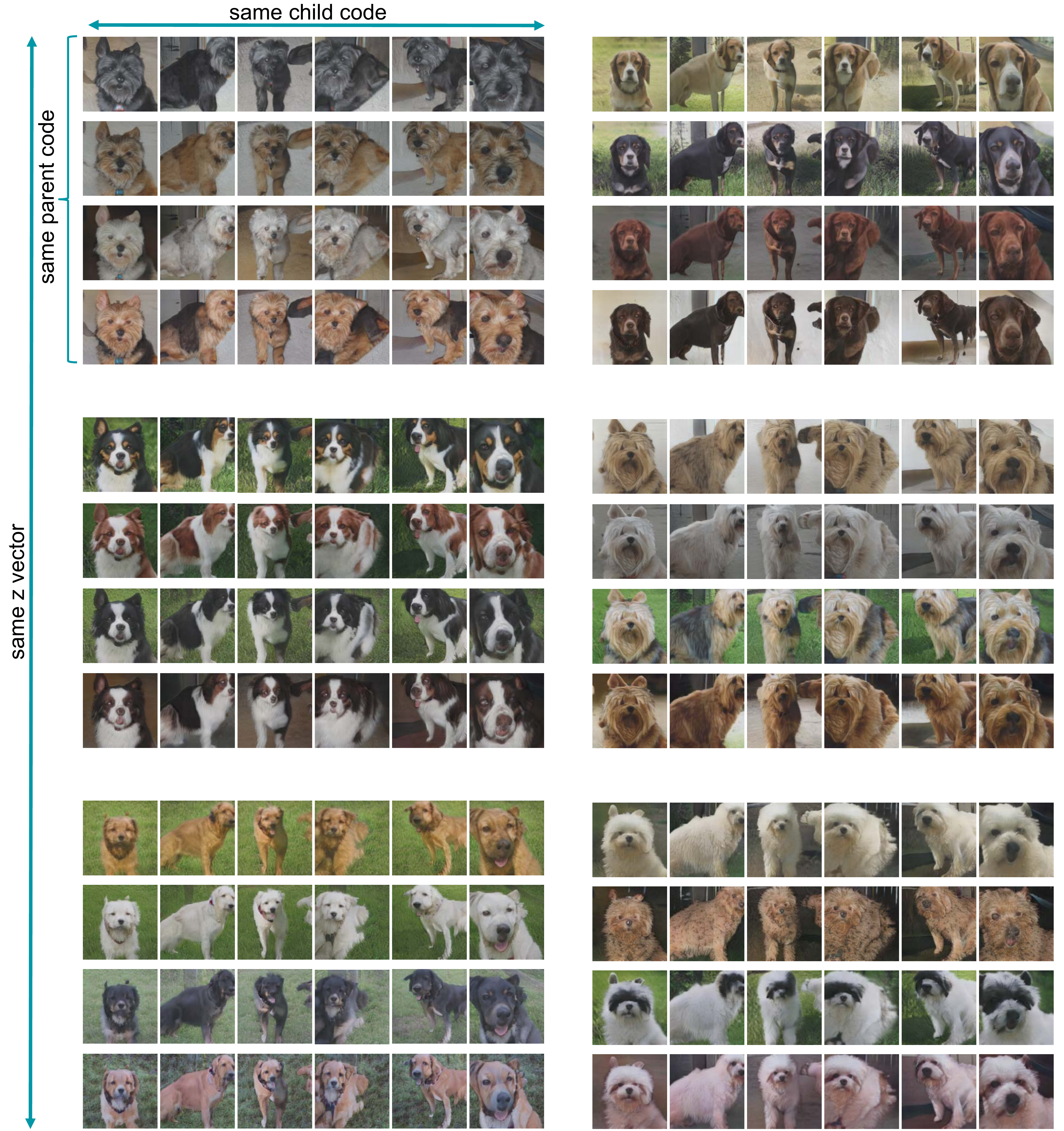}
    \caption{\textbf{Varying $p$ vs.~$c$ vs.~$z$ for dogs.} \emph{We show 6 different parent groups (with different $p$'s), each with 4 children (with different $c$'s).} For the same parent, the object's shape remains consistent while the appearance changes with different child codes. For the same child, the appearance remains consistent. Each column has the same random vector $z$ -- we can see that it controls the object's pose and position.}
    \label{fig:dogs_zc_vis}
\end{figure*}

\begin{figure*}[t!]
    \centering
    \includegraphics[width=0.9\textwidth]{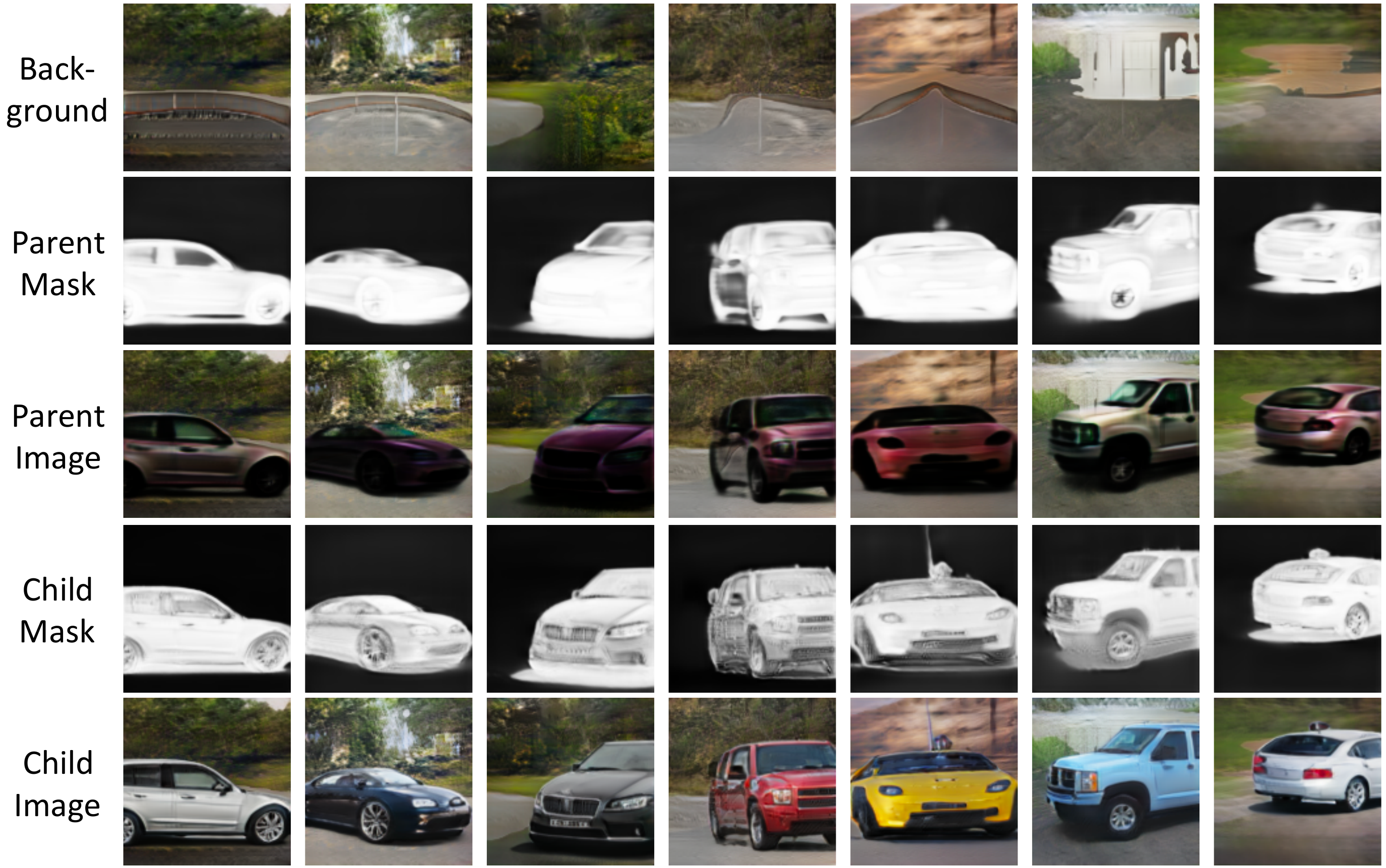}
    \caption{\textbf{FineGAN's stagewise image generation for cars.} Background stage generates a background which is retained over the child and parent stages. Parent stage generates a hollow image with only the object's shape, and child stage fills in the appearance.}
    \label{fig:all_stage_car}
\end{figure*}

\begin{figure*}[t!]
    \vspace{-10pt}
    \centering
    \includegraphics[width=1.0\textwidth]{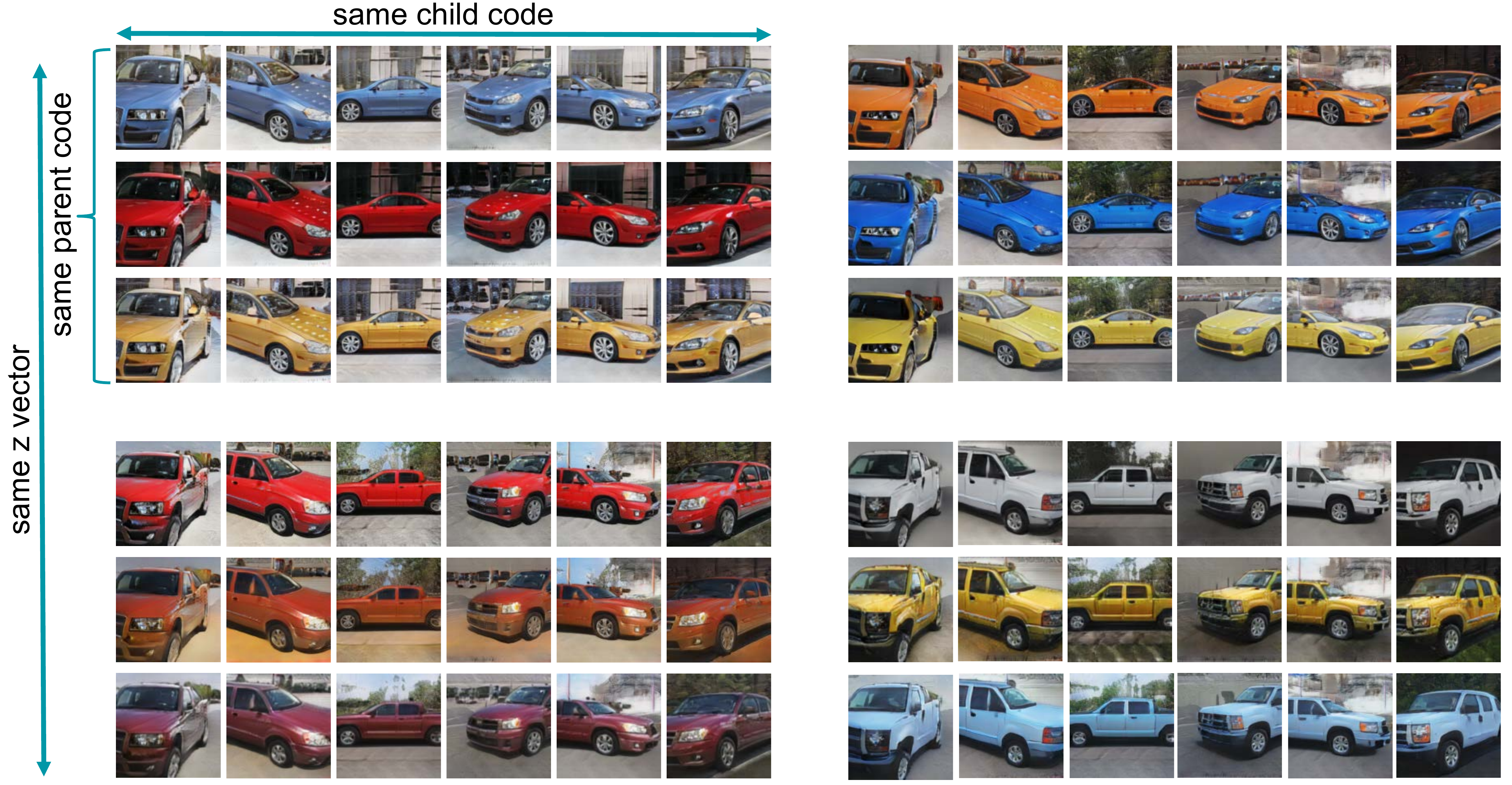}
    \caption{\textbf{Varying $p$ vs.~$c$ vs.~$z$ for cars.} \emph{We show 4 different parent groups (with different $p$'s), each with 3 children (with different $c$'s).} For the same parent, the object's shape remains consistent while the appearance changes with different child codes. For the same child, the appearance remains consistent. Each column has the same random vector $z$ -- we can see that it controls the object's pose and position.}
    \label{fig:cars_zc_vis}
\end{figure*}

Finally, we show additional qualitative results that supplement Figs.~3 and 4 in the main paper.  In Figs.~\ref{fig:all_stage_bird},~\ref{fig:all_stage_dog}, and~\ref{fig:all_stage_car}, we  show the stagewise image generation results for CUB, dogs, and cars respectively.  Figs.~\ref{fig:birds_zc_vis},~\ref{fig:dogs_zc_vis}, and~\ref{fig:cars_zc_vis} show the discovered grouping of parent and child codes by FineGAN for CUB, dogs, and cars respectively.

\section{Architecture and training details}

\subsection{FineGAN architecture}

\paragraph{Generative modules.} $G_b$ consists of six convolutional layers with BatchNorm~\cite{ioffe-icml15} and nearest neighbor upsampling applied after each convolutional layer except the last one. $G_p$ has an identical initial architecture as that of $G_b$. The intermediate feature representation obtained is concatenated with the parent code $p$, similar to StackGAN~\cite{stackgan2}. This representation is then passed through a residual block and a pair of convolutional layers, which gives us $F_p$ in Fig.2 of the main paper. $G_{p,f}$ and $G_{p,m}$ each consist of a single convolutional layer to transform the feature representation to have a resolution that matches the output image. Similar to the later part of $G_p$, $G_c$ consists of a residual block and a pair of convolutional layers which transform the concatenation of $F_p$ and child code $c$ into $F_c$. Each of $G_{c,f}$ and $G_{c,m}$, similar to $G_{p,f}$ and $G_{p,m}$, consist of a single convolutional layer.

\vspace{-10pt}
\paragraph{Discriminative modules.}  
$D_b$ consists of four convolutional layers, with leaky Relu used as the non-linearity except last convolutional layer, for which we use sigmoid non-linearity to output a $24$ x $24$ activation map of 0/1 (real/fake) scores. The network design is chosen such that each member from the $24$ x $24$ matrix represents the score for a $34$ x $34$ receptive field in the input image. We only consider background patches which lie completely outside the detected bounding box. $D_{aux}$ shares all convolutional layers with $D_b$ except the last one, where it branches out with a separate convolutional layer, again giving a $24$ x $24$ activation map of 0/1 scores indicating background/foreground classification. $D_p$ consists of eight convolutional layers, with all except last followed by BatchNorm and leaky Relu layers. This network outputs an $N_p$ dimensional parent class distribution. $D_c$ has an identical architecture as that of $D_p$, except it outputs a $N_c$ dimensional child class distribution. Similar to the background stage, $D_{adv}$ shares all convolutional layers except the last two, where it branches out using a separate convolutional layer which outputs a single 0/1 scalar value indicating the real/fake score for the final child image $\mathcal{C}$.

\subsection{Training details}
We optimize FineGAN using Adam with learning rate $2$ x $10^{-4}$, $\beta_1 = 0.5$, $\beta_2 = 0.999$. We use a minibatch of 16 images and train for 600 epochs. Following StackGAN~\cite{stackgan2}, we crop all images to 1.5$\times$ their available bounding boxes.

\vspace{-10pt}
\paragraph{Hard negative training}

Upon complete training of our $G/D$ models, we find that some noise vectors $\mathbf{z}$ = $\{z_1, z_2,\dots\}$ result in degenerate images. This property is characteristic of $z$, and not of any other latent code; i.e. $I = G(z, b, p, c)$ is a degenerate image $\forall$ $b \sim p_b, p \sim p_p, c \sim p_c$ if $z \in \mathbf{z}$. We also find that almost all of these degenerate images have very low scores predicted for the activated class corresponding to the input child code; i.e., $D_c(c|G_c(z, b, p, c)) \approx 0$.  We exploit this observation and continue training the $G/D$ models for a further $1$-$2$ epochs. This time we train alternatively on a normal batch of images and on a batch of degenerate images, which is formed using images $I_i = G_c(z_i, b, p, c)$ for which $D_c(c|I_i)$ values are low. More specifically, we choose $16$ (same as minibatch size) lowest scoring $I_i$ out of $160$ randomly generated $I_i$. This way of \textit{hard negative} training alleviates the problem of degenerate images to a great extent. Note that this is not something that can be applied to standard unconditional GANs as they do not produce class-conditional scores. Although the same technique can be used for InfoGAN, in our experiments, it did not result in any significant improvement.


\end{document}